%% file: main.tex
\documentclass[accepted]{uai2025} % after acceptance, for a revised version; 
\usepackage[american]{babel}
\usepackage{natbib} % has a nice set of citation styles and commands
\usepackage{mathtools} % amsmath with fixes and additions
\usepackage{booktabs} % commands to create good-looking tables
\usepackage{tikz} % nice language for creating drawings and diagrams
\usepackage{multirow}
\usepackage{amsfonts}
\usepackage{algorithm}
\usepackage{algpseudocode}

\newcommand{\diag}[1]{\operatorname{diag}(#1)}
\newcommand{\A}{\mathbf{A}}
\newcommand{\B}{\mathbf{B}}
\newcommand{\W}{\mathbf{W}}
\newcommand{\s}{\mathbf{s}}
\newcommand{\U}{\mathbf{U}}
\newcommand{\V}{\mathbf{V}}
\newcommand{\D}{\mathcal{D}}
\newcommand{\N}{\mathcal{N}}
\newcommand{\y}{\mathbf{y}}
\newcommand{\x}{\mathbf{x}}
\newcommand{\eps}{\boldsymbol{\epsilon}}
\newcommand{\btheta}{\boldsymbol{\theta}}
\newcommand{\R}{\mathbb{R}}
\newcommand{\z}{\mathbf{z}}
\newcommand{\bP}{\mathbf{P}}

\title{Scalable Bayesian Low-Rank Adaptation of Large Language Models via Stochastic Variational Subspace Inference}

\author{Colin Samplawski}
\author{Adam D. Cobb}
\author{Manoj Acharya}
\author{Ramneet Kaur}
\author{Susmit Jha}
\affil{%
    Neuro-Symbolic Computing and Intelligence Research Group\\
    Computer Science Laboratory\\
    SRI International
}
\begin{document}
\maketitle

\begin{abstract}
Despite their widespread use, large language models (LLMs) are known to hallucinate incorrect information and be poorly calibrated. This makes the uncertainty quantification of these models of critical importance, especially in high-stakes domains, such as autonomy and healthcare. Prior work has made Bayesian deep learning-based approaches to this problem more tractable by performing inference over the low-rank adaptation (LoRA) parameters of a fine-tuned model. While effective, these approaches struggle to scale to larger LLMs due to requiring further additional parameters compared to LoRA. In this work we present \textbf{Scala}ble \textbf{B}ayesian \textbf{L}ow-Rank Adaptation via Stochastic Variational Subspace Inference (ScalaBL). We perform Bayesian inference in an $r$-dimensional subspace, for LoRA rank $r$. By repurposing the LoRA parameters as projection matrices, we are able to map samples from this subspace into the full weight space of the LLM. This allows us to learn all the parameters of our approach using stochastic variational inference. Despite the low dimensionality of our subspace, we are able to achieve competitive performance with state-of-the-art approaches while only requiring  ${\sim}1000$ additional parameters. Furthermore, it allows us to scale up to the largest Bayesian LLM to date, with four times as a many base parameters as prior work. 
\end{abstract}

\input{intro}
\input{prelims}
\input{methods}

\input{experiments}

\input{conclusion}

\begin{acknowledgements} % will be removed in pdf 
This material is based upon work supported by the United States Air Force and DARPA under Contract No. FA8750-23-C-0519 and HR0011-24-9-0424, and the U.S. Army Research Laboratory under Cooperative Research Agreement W911NF-17-2-0196 and Defense Logistics Agency
(DLA) and the Advanced Research Projects Agency for Health (ARPA-H) under Contract Number
SP4701-23-C-0073. Any opinions, findings
and conclusions or recommendations expressed in this material are those of the author(s) and do not necessarily reflect the views of the United States Air Force, DARPA, the U.S. Army Research Laboratory, ARPA-H or the United States Government.
\end{acknowledgements}

\bibliography{sources}

\newpage
\onecolumn
\title{Appendix}
\maketitle
\input{appendix}

\end{document}

%% file: intro.tex
\begin{figure*}
    \centering
    \includegraphics[width=\linewidth]{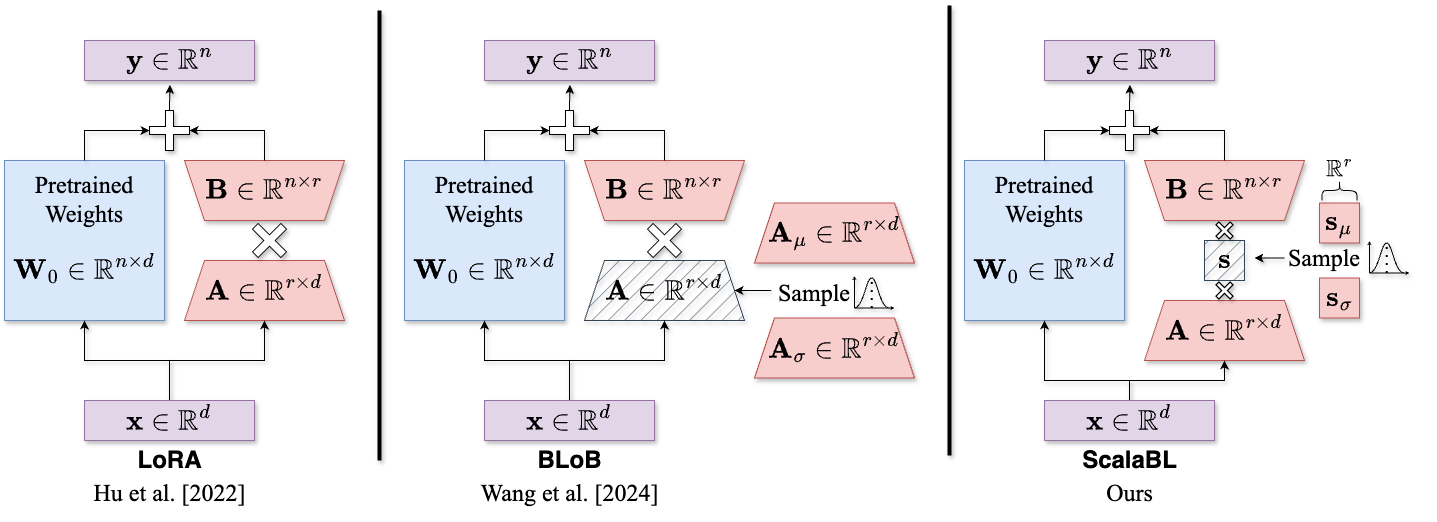}
    \caption{Visual depiction of prior work and our approach for a single layer. Blocks shaded red denote parameters which are trained and blue blocks denote frozen parameters. White blocks with hatching denote parameters which are sampled from learned variational distributions.}
    \label{fig:methods}
\end{figure*}

\section{Introduction}\label{sec:intro}
The use of large language models (LLMs) have become ubiquitous across many domains ranging from healthcare \citep{llmsmed}, scientific discovery \cite{zhang2024comprehensive}, cyber-physical systems \citep{aircraftverse}, code generation \citep{jiang2024survey}, and general everyday use \citep{anil2023gemini}. Therefore, ensuring that these models are reliable and trustworthy has never been more vital. However, it is well known that LLMs output incorrect information in the form of "hallucinations" \citep{huang2024survey} and are often poorly calibrated \citep{zhu2023calibration,spiess2024calibration}. One direction of research aimed at solving these issues considers quantifying the uncertainty of LLM outputs. A variety of post-hoc approaches have been proposed for this task, such as verbalized confidence \citep{tian2023just,xiongcan}, quantifying token level uncertainty \citep{sement,farquhar2024detecting}, or conformal prediction \citep{kaur2024addressing}.

In contrast, Bayesian deep learning (BDL) provides a principled approach to the uncertainty quantification of deep models. In this family of approaches, uncertainty quantification is performed by directly inferring a distribution over the weights of the model \citep{mcdropout,bbb,deepensembles}. Here we estimate a model's predictive uncertainty for a test instance $\mathbf{x}$, denoted $P(\y|\x, \D)$, by using Bayes' Rule to marginalize over the parameter posterior distribution, denoted $P(\W|\D)$, via the following integral:
\begin{align}
    P(\y|\x, \D) = \int P(\y|\x, \W) P(\W|\D) \mathrm{d} \W
\end{align}
where $\mathcal{D}$ is a training (or fine-tuning) dataset, and $\W$ are the model parameters. However, when scaling such techniques to LLMs, providing a good approximation of this intractable integral becomes increasingly challenging due to the large dimensionality of $\W$. For this reason, recent work has considered performing Bayesian inference over the smaller subset of parameters learned in popular parameter efficient fine-tuning (PEFT) approaches \citep{fu2023effectiveness}. 

In the widely-used low-rank adaptation (LoRA) technique of \cite{lora}, only a small subset of parameters are updated, saving considerable resources compared to updating the entire parameter set, while still enjoying most of the performance of the base model. Conveniently, the low dimensionality of these parameters additionally makes them well suited for BDL techniques. However, \cite{lap} and \cite{blob} have shown that directly applying BDL techniques such as Deep Ensembles \citep{deepensembles} or Monte Carlo Dropout \citep{mcdropout} over LoRA only leads to a marginal improvement on uncertainty quantification metrics compared to straightforward fine-tuning approaches such as maximum likelihood estimation (MLE) or Maximum a Posteriori (MAP). 

The first success in this space came from 
\citet{lap} who perform a Laplace approximation of the parameter posterior after MAP fine-tuning. 
The state-of-the-art approach of 
\citet{blob} instead uses stochastic variational inference in a technique they call Bayesian LoRA by Backprop (BLoB). Although this approach performs better than any previous approach, it comes at the cost of needing ${\sim}40\%$ more parameters than LoRA. This can be a major memory bottleneck in high-stakes, resource-constrained deployments where computing the Bayesian model average already stresses the available memory budget \citep{ursabench}. 

In this work, we introduce \textbf{Scala}ble \textbf{B}ayesian \textbf{L}ow Rank Adaptation via Stochastic Variational Subspace Inference (ScalaBL). As shown in Figure \ref{fig:methods}, we perform Bayesian inference inside a much smaller subspace of the full weight space $\W$ with dimensionality equal to the LoRA rank $r$. We show how we can repurpose the LoRA parameters $\A$ and $\B$ as projection matrices which map samples from the low dimensional subspace into the full weight space $\W$. We then learn the parameters of our approach using stochastic variational inference.

A major benefit of our approach is that it requires learning only $2r$ additional variational parameters for each LoRA layer, compared to the $rd$  parameters required by BLoB, where $d$ is the embedding dimension of the LLM. For example, when fine-tuning an LLM with 7 billion parameters where $d=3584$ using a rank of $r=8$, BLoB requires millions of additional parameters, while ScalaBL requires only ${\sim}1000$. Furthermore, so long as the rank $r$ remains constant, our approach requires the same number of additional parameters per layer regardless of the embedding dimension of the base LLM. As a result, we are able to scale our approach to a 32 billion base parameter model where $d=5120$, compared to the 7 billion parameter models considered by the prior work of \cite{lap} and \cite{blob}. Through extensive experimentation, we show that ScalaBL has competitive or superior performance compared to these state-of-the-art baselines on a suite of commonsense reasoning benchmarks in both in- and out-of-distribution settings.

We highlight our main contributions as follows:
\begin{itemize}[leftmargin=*, topsep=0pt, noitemsep]
    \item We propose ScalaBL, a Bayesian LoRA approach which performs stochastic variational inference inside a low dimensional subspace.
    \item ScalaBL enjoys considerable parameter efficiency compared to prior work and requires ${\sim}2000 \times$ fewer additional parameters.
    \item ScalaBL achieves competitive or superior performance to state-of-the-art approaches in terms of uncertainty quantification metrics, while requiring fewer parameters.
    \item Our work is the first to scale a Bayesian LoRA approach to a pre-trained model of 32 billion base parameters, compared to the 7 billion parameter models of prior work.
\end{itemize}

The structure of the paper is as follows.
In Section \ref{sec:prior_work}, we discuss relevant prior work that our approach builds on.
In Section \ref{sec:subspace}, we demonstrate our approach for building a parameter-efficient subspace and in Section \ref{sec:subspace_inference}, we discuss how to train a probabilistic model in this subspace using stochastic variational inference.
 In Section \ref{sec:experiments}, we provide results of our extensive experiments.
Finally, in Sections \ref{sec:limitations} and \ref{sec:conclusions}, we discuss limitations and conclude. Additional details and experimental results are included in the Appendix.
Our code is available at \url{github.com/SRI-CSL/BayesAdapt}. 

%% file: prelims.tex
\section{Prior Work}\label{sec:prior_work}
In this section we discuss prior work which our approach builds upon.

\subsection{Low-Rank Adaptation}
The low-rank adaptation (LoRA) approach of \cite{lora} has become a standard technique for fine-tuning LLMs in a tractable way. Consider a linear layer inside a pretrained LLM which has weights $\W_0 \in \R^{n \times d}$, where $d$ is the embedding dimension of the model and $n$ is the output dimension of the layer. A forward pass through the layer for a batch of $b$ input features $\mathbf{x} \in \mathbb{R}^{b \times d}$ is given by $\y = \x\W_0^T$.
In LoRA, rather than updating all of the model parameters, $\W_0$, we instead keep these parameters fixed and learn a new pair of low-rank parameters $\A \in \R^{r \times d}$ and $\B \in \R^{n \times r}$, such that:
\begin{align}
    \y = \x\W_0^T + \x (\B \A)^T
\end{align}
The value $r \ll \min(n,d)$ is commonly known as the LoRA rank. In this way, only $r(n+d)$ parameters need to be learned rather than $nd$, leading to considerable resource savings with minimal performance penalty.

 \subsection{Laplace LoRA}
The approach of \cite{lap} is the first example in the literature of applying uncertainty quantification techniques to LoRA layers by applying a Laplace approximation to the low-rank parameters. They treat a fine-tuned MAP estimate as the mean of a multivariate Gaussian distribution with covariance derived from the inverse Hessian. However, even when restricting the Laplace approximation to the LoRA parameters, evaluation of the Hessian is infeasible. Therefore \cite{lap} add structure to the Hessian by using a Kronecker factorization \citep{ritter2018scalable,daxberger2021laplace}. These Kronecker factors are still memory intensive, so \cite{lap} are forced to perform a further approximation via an iterative truncated singular value decomposition approach. The Laplace approximation is performed post-hoc after fine-tuning the LoRA parameters. An additional limitation is that at test time, they need to backpropagate through the model to build the approximated covariance matrix. This limits the scalability and use of their approach in resource-constrained environments.

\subsection{BLoB}
The current state-of-the-art approach in this space is Bayesian Low-Rank Adaptation by Backpropagation (BLoB).
BLoB moves away from the two stage approach of Laplace LoRA and instead performs stochastic variational inference over the LoRA parameters $\A$. More specifically, they follow the Bayes by Backprop  approach introduced by \cite{bbb}. That is, they recast $\mathbf{A}$ as the means of a low rank Gaussian distribution, denoted $\mathbf{A}_{\mu}$ and learn a set of variance parameters $\mathbf{A}_{\sigma}$. Using the reparameterization trick \citep{vae}, they project samples from this low rank distribution into the full weight space:
\begin{align}
    \W_t = \W_0 + \B (\A_{\mu}+\A_{\sigma} \cdot \eps_t)
\end{align}
where $\eps_t \sim \N(0,1)$.
\cite{blob} show empirically that their approach leads to better performance than Laplace LoRA. However, a notable upside of the Laplace approximation is that it does not require learning any additional parameters. Due to the variance parameters $\A_{\sigma}$, BLoB requires learning $1.4\times$ more parameters than the base LoRA fine-tuning process. Even for smaller 7 billion parameters models, this can be millions of additional parameters.

\subsection{Bayesian Subspace Inference}
Rather than trying to approximate the parameter posterior directly, \cite{izmailov2020subspace} purpose to perform Bayesian inference in a much smaller $k$-dimensional subspace of the full parameter space defined by vectors $\z \in \R^k$. They then learn a model $P(\z|\D)$, and a projection matrix $\bP$. This allows them to project samples into  full weight space:
\begin{align}
    \W_t = \W_0 + \bP \z_t \\
    \z_t \sim P(\z|\D)
\end{align}
As highlighted by \cite{izmailov2020subspace}, this model is not a reparameterization of the original parameter posterior as the projection into the full parameter space is not invertible. However, it has the upside that performing Bayesian inference in the subspace enables the use of many common Bayesian inference techniques that would otherwise be intractable. To the best of our knowledge, such subspace inference techniques have never been applied to LLMs.

%% file: methods.tex
\input{floats/algo}
\section{Methods}\label{sec:methods}
In this section we provide the details of our proposed approach. We first show how we construct an $r$-dimensional subspace of the full weight space $\W$ and then discuss how we train a probabilistic model in this subspace using stochastic variational inference.

\subsection{Subspace Construction}\label{sec:subspace}
To begin, consider a LoRA layer with rank $r$, initial weights $\W_0$, and low rank factors $\A$ and $\B$. We would like to generate an $r$-dimensional subspace defined by vectors $\s \in \R^{r}$ which can be projected into the full weight space $\W$. We retain $\B$ and use it as a projection matrix as in LoRA, allowing us to focus on building a subspace over $\A$. It is tempting to follow \cite{izmailov2020subspace} and learn a simple linear projection matrix $\bP$, resulting in the following subspace:
\begin{align}
    \mathcal{S}_{lin} = \{\W | \W = \W_0 +  \B \bP \s\}
\end{align}
However, $\A$ is an $r \times d$ matrix, so the product $\bP \s$ would need to be a vector of length $rd$ which then could be reshaped. This means that $\bP$ would need to have dimensionality $rd \times r$, and since we would take $\bP$ to be a matrix of learnable parameters, this choice of subspace would eliminate any parameter savings of LoRA. 

To motivate a more parameter efficient subspace construction, consider the truncated Singular Value Decomposition (SVD) of $\A$:
\begin{align}
    \U\diag{\s}\V = \text{SVD}(\A)
\end{align}
where $\s \in \R^r$ is the vector of singular values, $\U \in \R^{r \times r}, \V \in \R^{r \times d}$ are the left and right singular vectors, and $\diag{\cdot}$ is the diagonal embedding operator. Using $\U$ and $\V$ as projection matrices naturally defines the following subspace:
\begin{align}
    \mathcal{S}_{\text{SVD}} &= \{ \W | \W = \W_0 + \B \U \diag{\s} \V \} 
    \label{eq:svdsubspace}
\end{align}
We notice that $\B \U$ is a product of linear parameter matrices, which has the same representational power as $\B$ alone. Furthermore since the dimensionality of $\V$ is equal to that of $\A$, we simply rename $\V$ to be $\A$. This results in the following subspace:
\begin{align}
    \mathcal{S}_{\text{ScalaBL}} &= \{ \W | \W = \W_0 + \B \diag{\s} \A \} \\
    &= \{ \W | \W = \mathbf{f}(\s) \} 
\end{align}
where $\mathbf{f}$ is a projection function which is defined for notational convenance. 
Intuitively, we are repurposeing the LoRA parameters as projection matrices for an $r$-dimensional subspace that sits ``in-between'' $\A$ and $\B$.

\input{floats/tables/datasets}
\subsection{Variational Subspace Inference}\label{sec:subspace_inference}
Next we build a probabilistic model in this subspace with data likelihood given by:
\begin{align}
    P(\D | \s)= P(\D | \W=\mathbf{f}(\s) )
\end{align}
We set our variational approximation over $\s$ as an $r$-dimensional diagonal Gaussian distribution:
\begin{align}
    q_{\btheta}(\s)=\N(\s|\s_{\mu}, \diag{\s_{\sigma}})
\end{align}
with mean and variance parameters $\btheta=[\s_{\mu},\s_{\sigma}]$.
% Like BLoB, 
To learn these variational parameters we use stochastic variational inference. At training step $t$, we use the reparameterization trick \citep{vae} to generate a sample from $q_{\btheta}(\s)$ and project into the full weight space:
\begin{align}
    \W_t = \W_0 + \B \diag{\s_{\mu} + \s_{\sigma} \cdot \eps_t} \A
    \label{eq:reparam}
\end{align}
where $\eps_t \sim \N(0,1)$.
We then maximize the evidence lower bound (ELBO) \citep{elbo} for each batch $\D_t$:
\begin{align}
    \mathcal{L}_t &= \log P(\D_t|\W_t) - \beta D_{KL}( q_{\theta}(\s)|| P(\s))
\end{align}
Here the first term is the data likelihood under the LLM using the weight sample $\W_t$. The second term regularizes $q_{\theta}(\s)$ against a prior $P(\s)$, where $\beta$ is a scalar hyperparameter which controls the regularization strength \citep{betavae}. Our full training approach is shown in Algorithm \ref{alg:scalabl}. 

At test-time, we draw $N$ samples from $q_{\btheta}(\s)$, project them into the full weight space, and compute a Bayesian model average:
\begin{align}
    \mathbb{E}_{\s_n \sim q_{\btheta}(\s)} [ P(\y | \x, \s_n )] \approx \frac{1}{N} \sum_{n=1}^N  P(\y | \x, \mathbf{f}(\s_n))
\end{align}

We maintain $\A$ and $\B$ as learnable parameters as in LoRA. We then additionally need to learn just $2r$ variational parameters $\s_{\mu}$ and $\s_{\sigma}$. We note that BLoB can also be cast as a form of subspace inference where the LoRA layer itself defines the subspace of $\W$. This is a much higher dimensional subspace than the one used in ScalaBL, so performing variational inference results in needing to learn $rd$ additional parameters per LoRA layer.

%% file: floats/algo.tex
\begin{algorithm*}
\caption{ScalaBL Training Procedure}
\label{alg:scalabl}
\begin{algorithmic}[1]
\Require Pretrained weights $\W_0 \in \R^{n\times d}$, fine-tuning dataset $\D$, prior distribution $P(\s)$
\Require Number of training epochs $E$, batch size $B$, learning rate $\eta$
\Require KL divergence weight $\beta$, variance initialization parameter $\rho$
\State $\mathbf{Z} \sim \mathcal{U}(-\sqrt{\frac{1}{d}},\sqrt{\frac{1}{d}})$ \Comment{Sample random matrix}
\State $\textbf{\_}, \s_{\mu}, \A \gets \text{SVD}(\mathbf{Z})$  \Comment{Initialize using SVD}
\State $\B \gets 0$ \Comment{Initialize as in LoRA}
\State $\s_{\sigma} \sim \mathcal{U}(\frac{\rho}{\sqrt{2}}, \rho)$ \Comment{Initialize as in BLoB}
\For{epoch $e \gets 1 \dots E$}
\For{batch $\D_t \sim \D$}
\State $\eps_t \sim \N(0,1)$ \Comment{Sample noise}
\State $\W_t \gets \W_0 + \B \diag{\s_{\mu}+\s_{\sigma} \cdot \eps_t}\A $ \Comment{Reparameterization trick (Equation \ref{eq:reparam})}
\State $\mathcal{L}_t \gets -\frac{1}{B} \log P(\D_t|\W_t) + \beta D_{KL}(\N(\s_{\mu}, \diag{\s_{\sigma}})||P(\s))$ \Comment{Compute ELBO}
\State $\btheta \gets [\s_{\mu},\s_{\sigma,},\A,\B]$ \Comment{Collect trainable parameters}
\State $\btheta \gets \btheta - \eta \frac{\partial \mathcal{L}_t}{\partial \btheta}$\Comment{Compute gradient and update parameters}
\EndFor
\EndFor
\end{algorithmic}
\end{algorithm*}

%% file: floats/tables/datasets.tex
\begin{table*}[ht!]
\centering
\begin{tabular}{l|c|c|c|c}
\toprule
Dataset & Citation    & \# Classes    & Train Samples & Test Samples \\ \hline
Winogrande-Small (WG-S)       & \cite{winogrande} & 2& 0.64K & 1.27K\\
Winogrande-Medium (WG-M)      & \cite{winogrande} & 2&  2.56K & 1.27K\\
ARC-Easy (ARC-E)              & \cite{arc}        & 4&  2.25K & 0.57K\\
ARC-Challenge (ARC-C)         & \cite{arc}        & 4& 1.12K & 0.30K\\
OpenBookQA (OBQA)             & \cite{obqa}       & 4 & 4.96K & 0.50K \\
BoolQ                         & \cite{boolq}      & 2 &  2.49K & 3.27K\\
MMLU-Chemistry                    & \cite{mmlu}      & 4&  - & 0.10K\\
MMLU-Phyics                   & \cite{mmlu}      & 4&  - & 0.10K\\
\bottomrule
\end{tabular}
\caption{Commonsense Reasoning Datasets used in experiments. We note that the MMLU datasets are used only in the out-of-distribution experiments and therefore have no training samples.}
\label{tab:datasets}
\end{table*}

%% file: experiments.tex
\input{floats/tables/qwen7B_main}
\section{Experiments}\label{sec:experiments}
In this section we provide experimental comparisons between ScalaBL, several standard baselines, and current state-of-the-art approaches.

\subsection{Datasets}
Following the experimental protocol of \cite{lap} and \cite{blob} we fine-tune and evaluate our approach using  a suite of commonsense reasoning datasets shown in Table \ref{tab:datasets}. These datasets are posed as multiple choice questions. Given an input prompt with a question, we elicit the LLM's softmax distribution over the next token. We then select the logits for each possible answer (e.g. A,B,C,D) and renormalize. In this way, we transform these commonsense reasoning tasks into a classification task. This makes it straightforward to compute standard uncertainty metrics. In particular, we report classification accuracy (ACC), expected calibration error (ECE) \citep{guo2017calibration}, and the negative log likelihood (NLL) of the correct class. See Appendix Section \ref{sec:metric} for further details on these metrics.

\subsection{Baselines}
We compare against a suite of standard baselines. First we consider the standard LoRA training procedure with and without weight decay regularization (labeled MLE and MAP respectively). Next we compare against the standard BDL baselines of Deep Ensembles \citep{deepensembles}, and Monte Carlo Dropout \citep{mcdropout}. Finally, we present results against the two most recent state-of-the-art approaches: the Laplace approximation approach of \cite{lap} and BLoB \citep{blob}.

\subsection{Implementation Details}
We build our approach using the \texttt{bayesian-peft} library of \cite{blob}. This provides implementations of the standard baselines as well as BLoB. For the Laplace approximation we use the official code provided by \cite{lap}. In contrast to \cite{lap} and \cite{blob}, we present results on the newer \texttt{Qwen2.5} \citep{yang2024qwen2} family of models, rather than the older \texttt{Llama-2-7b} model \citep{llama2} of prior work. For the sake of comparison, results using \texttt{Llama-2-7b} are provided in Appendix Section \ref{sec:llama2}. 

 Following \cite{lap} and \cite{blob}, we apply LoRA to the query and value parameters of each self-attention layer as well as the softmax output head of the LLM using rank of $r=8$. We follow the training procedure and hyperparameters of BLoB.
 All approaches are trained for 5000 steps using the AdamW optimizer. Training was performed using a batch size of 4 for the 7 billion parameter models and a batch size of 2 for the 32 billion parameter model. In contrast to \cite{blob}, we train all approaches using 16-bit precision for the frozen model parameters instead of using 8-bit quantization. The learnable model parameters remain 32-bit. All experiments were performed on a single 80GB NVIDIA A100 GPU.

For ScalaBL, we use the same KL weighting schedule as BLoB with an maximum value of $\beta=0.1$ We do not use the Flipout technique \citep{wen2018flipout} that was utilized by BLoB as we found that it did not noticeably effect performance. This simplifies the complexity of the implementation of our approach compared to BLoB. As in BLoB, we use a standard $\mathcal{N}(0,I_r)$ as the prior $P(\s)$. We initialize $\s_{\mu}$ and $\A$ by performing an SVD on a randomly initialized matrix. This is a fast operation due to the low rank nature of the LoRA matrices. Like in BLoB, the variance parameters $\s_{\sigma}$ were initialized as small uniformly random values. We use a log parametrization for $\s_{\sigma}$ 
to ensure the variance remain positive. Following the intuition that $\s_{\mu}$ is analogous to the singular values of $\A$, we ensure their positivity using a log parametrization as well.

For the variational approaches, BLoB and ScalaBL, we present our main results using $N=10$ posterior weight samples during evaluation, which \cite{blob} found to give the best performance. The effect of this hyperparameter is explored further in Appendix Section \ref{sec:samples}. Similarly, we perform 10 forward passes for the MC-Dropout baseline. For Deep Ensembles, we use an ensemble size of 3.

\input{floats/tables/qwen7B_ood}
\subsection{In-Distribution Results}
In Table \ref{tab:qwen7B_main} we present test set results for a standard in-distribution setting using the \texttt{Qwen2.5-7B} LLM. We first notice that a straightforward MLE fine-tuning approach leads to high accuracy across all datasets, but often overfits as evidenced by the poor ECE results. The MAP result is equivalent to MLE with a weight decay penalty of $10^{-2}$, which marginally improves final calibration. We see minor improvements in ECE and NLL when moving to Monte Carlo Dropout and Deep Ensembles, with Deep Ensembles performing the best of the standard baselines, albeit at significantly higher resource cost.

Validating the results of \cite{lap} and \cite{blob}, we see that the more recent state-of-the-art approaches out perform the baselines in terms of ECE and NLL, with minimal reduction in classification accuracy. Furthermore, we see that ScalaBL consistently achieves performance that is competitive with BLoB, and even achieves state-of-the-art performance on  Winogrande-Medium dataset in terms of ECE. 

Unsurprisingly, BLoB often performs the best out of all methods and regularly outperforms ScalaBL by a small margin. However, BLoB has strictly greater representational power than ScalaBL or Laplace due to its higher parameter count. Compared to MLE, BLoB requires an additional ${\sim}1.4\times$ as many parameters, while ScalaBL requires only ${\sim}1.0001\times$ as many. For this choice of LLM and rank, this results in BLoB adding ${\sim}1.6$ million parameters on top of MLE, while ScalaBL adds only 912. With that in mind, ScalaBL achieves very competitive performance at much lower cost compared to BLoB. For example, on the ARC-Challenge dataset, BLoB  sees ${\sim}1.3\times$ better ECE performance than ScalaBL with similar accuracy. However, BLoB requires $1792 \times$ more additional parameters than ScalaBL.

\subsection{Out-of-Distribution Results}
Next we consider an out-of-distribution experiment where models are trained on the OpenBookQA (OBQA) dataset which consists of grade school level, multiple choice science questions. First we evaluate this tuned model on the ARC datasets, which also consists of grade school level multiple choices, representing a smaller distribution shift. Next we investigate a larger distribution shift by evaluating on the more challenging MMLU-Chemistry and MMLU-Physics datasets which consist of undergraduate level chemistry and physics multiple choice questions, respectively. The results of this experiment for all methods are displayed in Table \ref{tab:qwen7B_ood}.

We again notice that the recent state-of-the-art approaches outperform the standard baselines in terms of uncertainty quantification with comparable accuracy. We see that all methods experience worse calibration when tested under large distribution shift. We additionally point out the poor accuracy of the Laplace method on the MMLU datasets. We see strong performance of our proposed method, with ScalaBL out competing BLoB and Laplace on several datasets in terms of ECE. Under a both small and large amounts of distribution shift, ScalaBL achieves comparable performance to BLoB across all metrics.

\input{floats/tables/qwen32B_main}
\subsection{Scaling to Larger Models}\label{sec:larger_models}
A limitation of the prior work of \cite{lap} and \cite{blob} is their use of relatively small LLMs with only 7 billion parameters. This makes it unclear if their experimental conclusions generalize to the much larger model sizes which are currently in use \citep{anil2023gemini}. For this reason we consider scaling our approach to the largest Bayesian LLM to date, \texttt{Qwen2.5-32B}, with four times as many base parameters as prior work. We conduct the same in-distribution experiments as before and present test set results in Table \ref{tab:qwen32B_main}. We note that we do not report results using the Laplace baseline as its post-hoc procedure exceeded the memory availability of our 80GB A100 GPU even when using 8-bit parameters and test time batch size of 1, underscoring the poor scalability of this method.

In contrast to earlier results, standard baselines are much more competitive when using a larger base model. We see that even simple techniques, such as MLE or MAP, lead to models which are much better calibrated than their smaller counterparts. This phenomenon has been noticed in prior work \citep{xiongcan,spiess2024calibration}. Furthermore, we see that Deep Ensembles is often the best performing approach across all three metrics. However, this comes at significantly higher resource usage. 

We observe that our proposed approach, ScalaBL, continues to show competitive performance against the baselines, including BLoB. It often performs the second best in terms of ECE and NLL, while experiencing a similar classification performance as BLoB. When using a larger base model, the efficiency and scalability of our method is even more pronounced. Moving from \texttt{Qwen2.5-7B} to \texttt{Qwen2.5-32B} increases the model's embedding dimension from 3584 to 5120 and adds an additional 12 layers. Since the number of variance parameters in BLoB scales with the embedding dimension, it now requires an additional ${\sim}5.2$ million parameters compared to MLE. By contrast ScalaBL's additional parameter count scales only with $r$ which was not changed for this larger model. For this reason, ScalaBL only requires adding an additional 2064 parameters. In fact, for this choice of LLM and rank, BLoB requires adding $2560\times$ more parameters than ScalaBL for similar performance. For this reason we feel that ScalaBL is the only method that is capable of scaling to current frontier models, which are already over a trillion base parameters \cite{anil2023gemini}.

%% file: floats/tables/qwen7B_main.tex
\begin{table*}[h!]
\centering
\caption{In-distribution experiment using \texttt{Qwen2.5-7B}. 
We report the mean and standard deviation of test set performance using 8 training seeds.
\textbf{Bold} and \underline{underlined} results denote the best and second best mean performance on each metric/dataset.}
\begin{tabular}{@{}ccccccccc@{}}
\toprule
\textbf{Metric} & \textbf{Method} & \textbf{Params (M)} & \textbf{WG-S} & \textbf{ARC-C} & \textbf{ARC-E} & \textbf{WG-M} & \textbf{OBQA} & \textbf{BoolQ} \\
\midrule
\multirow{6}{*}{\textbf{ACC ($\uparrow$)}}
& MLE & $3.768$ & $78.86_{\pm 0.8}$ & $\underline{89.53}_{\pm 0.4}$ & $95.60_{\pm 0.2}$ & $82.30_{\pm 0.7}$ & $\underline{92.25}_{\pm 0.9}$ & $\underline{89.06}_{\pm 0.2}$ \\
& MAP & $3.768$ & $\underline{78.94}_{\pm 0.8}$ & $88.98_{\pm 0.8}$ & $95.73_{\pm 0.3}$ & $82.09_{\pm 0.7}$ & $91.72_{\pm 0.7}$ & $89.04_{\pm 0.1}$ \\
& MC-Dropout & $3.768$ & $78.57_{\pm 0.4}$ & $89.44_{\pm 0.3}$ & $95.77_{\pm 0.4}$ & $\underline{82.72}_{\pm 1.0}$ & $91.80_{\pm 0.6}$ & $88.99_{\pm 0.1}$ \\
& Ensemble & $11.305$ & $\textbf{79.09}_{\pm 0.4}$ & $89.44_{\pm 0.5}$ & $95.73_{\pm 0.1}$ & $\textbf{83.23}_{\pm 0.4}$ & $\textbf{92.70}_{\pm 0.6}$ & $\textbf{89.13}_{\pm 0.1}$ \\
& Laplace & $3.768$ & $77.28_{\pm 0.6}$ & $85.25_{\pm 1.3}$ & $95.34_{\pm 0.5}$ & $81.99_{\pm 0.7}$ & $91.68_{\pm 0.4}$ & $87.77_{\pm 0.1}$ \\
& BLoB & $5.403$ & $78.66_{\pm 0.7}$ & $\underline{89.53}_{\pm 0.8}$ & $\textbf{96.54}_{\pm 0.3}$ & $82.30_{\pm 0.3}$ & $91.72_{\pm 0.7}$ & $89.05_{\pm 0.2}$ \\
& ScalaBL (ours) & $3.769$ & $78.64_{\pm 0.4}$ & $\textbf{90.16}_{\pm 0.8}$ & $\underline{96.26}_{\pm 0.1}$ & $81.42_{\pm 0.3}$ & $90.90_{\pm 0.5}$ & $88.48_{\pm 0.1}$ \\
\midrule
\multirow{6}{*}{\textbf{ECE ($\downarrow$)}}
& MLE & $3.768$ & $20.14_{\pm 0.9}$ & $10.11_{\pm 0.5}$ & $4.17_{\pm 0.2}$ & $16.10_{\pm 0.6}$ & $6.40_{\pm 0.8}$ & $3.79_{\pm 0.1}$ \\
& MAP & $3.768$ & $19.99_{\pm 0.9}$ & $10.54_{\pm 0.7}$ & $4.08_{\pm 0.2}$ & $16.42_{\pm 0.8}$ & $6.61_{\pm 0.6}$ & $3.81_{\pm 0.1}$ \\
& MC-Dropout & $3.768$ & $20.15_{\pm 0.3}$ & $10.06_{\pm 0.4}$ & $4.01_{\pm 0.3}$ & $15.46_{\pm 0.8}$ & $6.60_{\pm 0.4}$ & $3.88_{\pm 0.1}$ \\
& Ensemble & $11.305$ & $19.06_{\pm 0.4}$ & $10.13_{\pm 0.4}$ & $3.75_{\pm 0.1}$ & $13.65_{\pm 0.8}$ & $4.96_{\pm 0.6}$ & $2.61_{\pm 0.1}$ \\
& Laplace & $3.768$ & $13.32_{\pm 3.6}$ & $37.90_{\pm 2.1}$ & $33.80_{\pm 3.8}$ & $\underline{4.81}_{\pm 0.7}$ & $\textbf{1.90}_{\pm 0.4}$ & $\textbf{1.18}_{\pm 0.2}$ \\
& BLoB & $5.403$ & $\textbf{7.88}_{\pm 0.3}$ & $\textbf{4.03}_{\pm 1.0}$ & $\textbf{1.60}_{\pm 0.4}$ & $5.08_{\pm 0.4}$ & $\underline{2.16}_{\pm 0.5}$ & $\underline{1.40}_{\pm 0.3}$ \\
& ScalaBL (ours) & $3.769$ & $\underline{8.88}_{\pm 0.5}$ & $\underline{5.03}_{\pm 0.9}$ & $\underline{1.78}_{\pm 0.2}$ & $\textbf{3.64}_{\pm 0.2}$ & $2.43_{\pm 0.7}$ & $1.96_{\pm 0.3}$ \\
\midrule
\multirow{6}{*}{\textbf{NLL ($\downarrow$)}}
& MLE & $3.768$ & $1.94_{\pm 0.3}$ & $1.05_{\pm 0.1}$ & $0.44_{\pm 0.0}$ & $1.20_{\pm 0.1}$ & $0.38_{\pm 0.1}$ & $0.25_{\pm 0.0}$ \\
& MAP & $3.768$ & $1.88_{\pm 0.2}$ & $1.05_{\pm 0.1}$ & $0.43_{\pm 0.0}$ & $1.27_{\pm 0.2}$ & $0.39_{\pm 0.0}$ & $0.25_{\pm 0.0}$ \\
& MC-Dropout & $3.768$ & $1.90_{\pm 0.2}$ & $1.02_{\pm 0.1}$ & $0.43_{\pm 0.0}$ & $1.07_{\pm 0.0}$ & $0.36_{\pm 0.0}$ & $0.25_{\pm 0.0}$ \\
& Ensemble & $11.305$ & $1.33_{\pm 0.1}$ & $0.75_{\pm 0.0}$ & $0.25_{\pm 0.0}$ & $0.74_{\pm 0.0}$ & $0.27_{\pm 0.0}$ & $\underline{0.24}_{\pm 0.0}$ \\
& Laplace & $3.768$ & $\underline{0.55}_{\pm 0.0}$ & $0.80_{\pm 0.1}$ & $0.51_{\pm 0.1}$ & $0.44_{\pm 0.0}$ & $\underline{0.23}_{\pm 0.0}$ & $0.29_{\pm 0.0}$ \\
& BLoB & $5.403$ & $\textbf{0.51}_{\pm 0.0}$ & $\textbf{0.30}_{\pm 0.0}$ & $\textbf{0.10}_{\pm 0.0}$ & $\textbf{0.39}_{\pm 0.0}$ & $\textbf{0.21}_{\pm 0.0}$ & $\textbf{0.23}_{\pm 0.0}$ \\
& ScalaBL (ours) & $3.769$ & $\textbf{0.51}_{\pm 0.0}$ & $\underline{0.31}_{\pm 0.0}$ & $\underline{0.11}_{\pm 0.0}$ & $\underline{0.40}_{\pm 0.0}$ & $\underline{0.23}_{\pm 0.0}$ & $\underline{0.24}_{\pm 0.0}$ \\
\bottomrule
\end{tabular}
\label{tab:qwen7B_main}
\end{table*}

%% file: floats/tables/qwen7B_ood.tex
\begin{table*}[h!]
\centering
\caption{Out-of-distribution experiment using \texttt{Qwen2.5-7B}. 
We report the mean and standard deviation of test set performance using 8 training seeds.
\textbf{Bold} and \underline{underlined} results denote the best and second best mean performance on each metric/dataset.
}\begin{tabular}{@{}cccccccc@{}}
\toprule
\textbf{Metric} & \textbf{Method} & \textbf{Params (M)} & \textbf{OBQA} & \textbf{ARC-C} & \textbf{ARC-E} & \textbf{Chemistry} & \textbf{Physics} \\
\midrule
\multirow{6}{*}{\textbf{ACC ($\uparrow$)}}
& MLE & $3.768$ & $\underline{92.25}_{\pm 0.9}$ & $90.88_{\pm 0.7}$ & $95.64_{\pm 0.5}$ & $53.00_{\pm 1.3}$ & $53.00_{\pm 1.5}$ \\
& MAP & $3.768$ & $91.72_{\pm 0.7}$ & $90.20_{\pm 0.9}$ & $95.53_{\pm 0.6}$ & $\underline{53.50}_{\pm 0.9}$ & $53.25_{\pm 3.1}$ \\
& MC-Dropout & $3.768$ & $91.80_{\pm 0.6}$ & $90.37_{\pm 0.5}$ & $95.51_{\pm 0.4}$ & $52.75_{\pm 1.3}$ & $51.00_{\pm 2.1}$ \\
& Ensemble & $11.305$ & $\textbf{92.70}_{\pm 0.6}$ & $90.84_{\pm 0.6}$ & $95.71_{\pm 0.5}$ & $53.25_{\pm 1.0}$ & $\textbf{53.88}_{\pm 1.2}$ \\
& Laplace & $3.768$ & $91.68_{\pm 0.4}$ & $90.51_{\pm 0.7}$ & $95.61_{\pm 0.4}$ & $48.75_{\pm 1.8}$ & $50.74_{\pm 2.3}$ \\
& BLoB & $5.403$ & $91.72_{\pm 0.7}$ & $\textbf{92.49}_{\pm 0.5}$ & $\textbf{96.07}_{\pm 0.5}$ & $\textbf{54.69}_{\pm 1.4}$ & $\underline{53.65}_{\pm 2.8}$ \\
& ScalaBL (ours) & $3.769$ & $90.90_{\pm 0.5}$ & $\underline{91.06}_{\pm 1.1}$ & $\underline{95.74}_{\pm 0.5}$ & $52.60_{\pm 1.8}$ & $53.13_{\pm 1.5}$ \\
\midrule
\multirow{6}{*}{\textbf{ECE ($\downarrow$)}}
& MLE & $3.768$ & $6.40_{\pm 0.8}$ & $7.72_{\pm 0.6}$ & $3.48_{\pm 0.4}$ & $23.29_{\pm 2.2}$ & $23.22_{\pm 3.2}$ \\
& MAP & $3.768$ & $6.61_{\pm 0.6}$ & $7.89_{\pm 0.9}$ & $3.31_{\pm 0.2}$ & $22.90_{\pm 1.9}$ & $21.52_{\pm 4.4}$ \\
& MC-Dropout & $3.768$ & $6.60_{\pm 0.4}$ & $7.63_{\pm 0.8}$ & $3.38_{\pm 0.2}$ & $23.74_{\pm 1.6}$ & $21.61_{\pm 2.1}$ \\
& Ensemble & $11.305$ & $4.96_{\pm 0.6}$ & $6.18_{\pm 0.6}$ & $2.63_{\pm 0.4}$ & $19.49_{\pm 1.4}$ & $17.33_{\pm 2.0}$ \\
& Laplace & $3.768$ & $\textbf{1.90}_{\pm 0.4}$ & $4.75_{\pm 0.7}$ & $\textbf{1.99}_{\pm 0.4}$ & $\textbf{14.31}_{\pm 2.1}$ & $\textbf{11.94}_{\pm 4.5}$ \\
& BLoB & $5.403$ & $\underline{2.16}_{\pm 0.5}$ & $\underline{4.46}_{\pm 0.5}$ & $2.35_{\pm 0.4}$ & $\underline{16.21}_{\pm 2.2}$ & $16.93_{\pm 2.4}$ \\
& ScalaBL (ours) & $3.769$ & $2.43_{\pm 0.7}$ & $\textbf{4.41}_{\pm 0.7}$ & $\underline{1.92}_{\pm 0.4}$ & $16.94_{\pm 1.8}$ & $\underline{16.29}_{\pm 1.8}$ \\
\midrule
\multirow{6}{*}{\textbf{NLL ($\downarrow$)}}
& MLE & $3.768$ & $0.38_{\pm 0.1}$ & $0.44_{\pm 0.0}$ & $0.23_{\pm 0.0}$ & $1.53_{\pm 0.1}$ & $1.18_{\pm 0.1}$ \\
& MAP & $3.768$ & $0.39_{\pm 0.0}$ & $0.46_{\pm 0.0}$ & $0.22_{\pm 0.0}$ & $1.52_{\pm 0.1}$ & $1.19_{\pm 0.1}$ \\
& MC-Dropout & $3.768$ & $0.36_{\pm 0.0}$ & $0.43_{\pm 0.0}$ & $0.21_{\pm 0.0}$ & $1.50_{\pm 0.1}$ & $1.19_{\pm 0.0}$ \\
& Ensemble & $11.305$ & $0.27_{\pm 0.0}$ & $0.33_{\pm 0.0}$ & $0.17_{\pm 0.0}$ & $1.29_{\pm 0.0}$ & $1.07_{\pm 0.0}$ \\
& Laplace & $3.768$ & $0.23_{\pm 0.0}$ & $0.32_{\pm 0.0}$ & $\underline{0.15}_{\pm 0.0}$ & $\textbf{1.11}_{\pm 0.0}$ & $1.03_{\pm 0.0}$ \\
& BLoB & $5.403$ & $\textbf{0.21}_{\pm 0.0}$ & $\underline{0.28}_{\pm 0.0}$ & $\underline{0.15}_{\pm 0.0}$ & $1.32_{\pm 0.1}$ & $\underline{0.99}_{\pm 0.0}$ \\
& ScalaBL (ours) & $3.769$ & $\underline{0.23}_{\pm 0.0}$ & $\textbf{0.27}_{\pm 0.0}$ & $\textbf{0.14}_{\pm 0.0}$ & $\underline{1.26}_{\pm 0.0}$ & $\textbf{0.96}_{\pm 0.0}$ \\
\bottomrule
\end{tabular}
\label{tab:qwen7B_ood}
\end{table*}

%% file: floats/tables/qwen32B_main.tex
\begin{table*}
\centering
\caption{In-distribution experiment using \texttt{Qwen2.5-32B}. 
We report the mean and standard deviation of test set performance using 3 training seeds.
\textbf{Bold} and \underline{underlined} results denote the best and second best mean performance on each metric/dataset. %\textcolor{RK}{Here also a line on Ensemble achieves SOTA results but at the cost of much higher (maybe a no., if we know) resource usage than ScalaBL, which achieves competetive results.}
}
\begin{tabular}{@{}ccc|cccccc@{}}
\toprule
\multirow{2}{*}{\textbf{Metric}} & \multirow{2}{*}{\textbf{Method}}  & \multirow{2}{*}{\textbf{Params (M)}} & \multicolumn{6}{c}{\textbf{Datasets}} \\ 
  &  & & \textbf{WG-S} & \textbf{ARC-C} & \textbf{ARC-E} & \textbf{WG-M} & \textbf{OBQA} & \textbf{BoolQ} \\
\midrule
\multirow{6}{*}{\textbf{ACC ($\uparrow$)}}
& MLE & $9.646$ & $86.45_{\pm 0.6}$ & $93.90_{\pm 0.9}$ & $\underline{98.80}_{\pm 0.3}$ & $90.90_{\pm 0.3}$ & $\textbf{96.93}_{\pm 0.6}$ & $\underline{91.42}_{\pm 0.1}$ \\
& MAP & $9.646$ & $86.73_{\pm 0.8}$ & $93.85_{\pm 1.0}$ & $\textbf{98.83}_{\pm 0.4}$ & $\underline{91.00}_{\pm 0.1}$ & $\textbf{96.93}_{\pm 0.7}$ & $\textbf{91.47}_{\pm 0.1}$ \\
& MC-Dropout & $9.646$ & $\underline{86.81}_{\pm 0.5}$ & $94.18_{\pm 0.4}$ & $98.71_{\pm 0.5}$ & $90.63_{\pm 0.7}$ & $\underline{96.60}_{\pm 0.7}$ & $\underline{91.42}_{\pm 0.0}$ \\
& Ensemble & $28.938$ & $\textbf{86.99}_{\pm 0.4}$ & $\textbf{94.97}_{\pm 0.3}$ & $98.65_{\pm 0.3}$ & $\textbf{91.42}_{\pm 0.2}$ & $\textbf{96.93}_{\pm 0.6}$ & $91.11_{\pm 0.1}$ \\
& BLoB & $14.930$ & $84.92_{\pm 0.4}$ & $\underline{94.07}_{\pm 0.4}$ & $98.65_{\pm 0.5}$ & $90.71_{\pm 0.3}$ & $96.53_{\pm 0.3}$ & $90.57_{\pm 0.2}$ \\
%& ScalaBL (SVD) & $9.656$ & $84.48_{\pm 0.3}$ & $92.95_{\pm 0.0}$ & $98.65_{\pm 0.1}$ & $90.42_{\pm 0.3}$ & $\underline{96.60}_{\pm 0.4}$ & $91.03_{\pm 0.0}$ \\
& ScalaBL (ours) & $9.648$ & $84.73_{\pm 0.4}$ & $93.74_{\pm 0.7}$ & $98.65_{\pm 0.2}$ & $90.07_{\pm 0.1}$ & $96.33_{\pm 0.2}$ & $90.99_{\pm 0.1}$ \\
\midrule
\multirow{6}{*}{\textbf{ECE ($\downarrow$)}}
& MLE & $9.646$ & $12.85_{\pm 0.7}$ & $5.88_{\pm 0.8}$ & $1.04_{\pm 0.3}$ & $7.11_{\pm 0.4}$ & $2.18_{\pm 0.4}$ & $1.66_{\pm 0.1}$ \\
& MAP & $9.646$ & $12.48_{\pm 0.9}$ & $5.97_{\pm 0.7}$ & $\underline{0.96}_{\pm 0.4}$ & $6.83_{\pm 0.4}$ & $\underline{2.03}_{\pm 0.6}$ & $1.66_{\pm 0.2}$ \\
& MC-Dropout & $9.646$ & $12.22_{\pm 0.5}$ & $5.38_{\pm 0.3}$ & $1.22_{\pm 0.2}$ & $7.50_{\pm 0.2}$ & $2.50_{\pm 0.3}$ & $1.50_{\pm 0.1}$ \\
& Ensemble & $28.938$ & $11.20_{\pm 0.6}$ & $\textbf{4.89}_{\pm 0.2}$ & $\textbf{0.98}_{\pm 0.4}$ & $\textbf{5.02}_{\pm 0.1}$ & $\textbf{1.85}_{\pm 0.3}$ & $\textbf{0.74}_{\pm 0.1}$ \\
& BLoB & $14.930$ & $\textbf{7.49}_{\pm 0.3}$ & $5.07_{\pm 0.3}$ & $1.11_{\pm 0.3}$ & $6.18_{\pm 0.5}$ & $2.51_{\pm 0.5}$ & $\underline{1.39}_{\pm 0.1}$ \\
%& ScalaBL (SVD) & $9.656$ & $\underline{9.86}_{\pm 0.7}$ & $5.63_{\pm 0.4}$ & $1.12_{\pm 0.2}$ & $\underline{5.44}_{\pm 0.3}$ & $2.44_{\pm 0.4}$ & $1.47_{\pm 0.1}$ \\
& ScalaBL (ours) & $9.648$ & $\underline{10.92}_{\pm 0.3}$ & $\underline{5.03}_{\pm 0.6}$ & $1.06_{\pm 0.1}$ & $\underline{5.91}_{\pm 0.2}$ & $2.32_{\pm 0.6}$ & $1.40_{\pm 0.2}$ \\
\midrule
\multirow{6}{*}{\textbf{NLL ($\downarrow$)}}
& MLE & $9.646$ & $1.08_{\pm 0.1}$ & $0.49_{\pm 0.1}$ & $0.06_{\pm 0.0}$ & $0.35_{\pm 0.0}$ & $0.13_{\pm 0.0}$ & $\underline{0.18}_{\pm 0.0}$ \\
& MAP & $9.646$ & $1.05_{\pm 0.0}$ & $0.53_{\pm 0.0}$ & $0.06_{\pm 0.0}$ & $0.33_{\pm 0.0}$ & $0.13_{\pm 0.0}$ & $\underline{0.18}_{\pm 0.0}$ \\
& MC-Dropout & $9.646$ & $0.99_{\pm 0.0}$ & $0.50_{\pm 0.0}$ & $0.07_{\pm 0.0}$ & $0.36_{\pm 0.0}$ & $0.14_{\pm 0.0}$ & $\underline{0.18}_{\pm 0.0}$ \\
& Ensemble & $28.938$ & $0.67_{\pm 0.0}$ & $\textbf{0.30}_{\pm 0.0}$ & $\textbf{0.04}_{\pm 0.0}$ & $\textbf{0.25}_{\pm 0.0}$ & $\textbf{0.11}_{\pm 0.0}$ & $\underline{0.18}_{\pm 0.0}$ \\
& BLoB & $14.930$ & $\textbf{0.44}_{\pm 0.0}$ & $0.40_{\pm 0.0}$ & $0.06_{\pm 0.0}$ & $\underline{0.30}_{\pm 0.0}$ & $\underline{0.12}_{\pm 0.0}$ & $\textbf{0.17}_{\pm 0.0}$ \\
%& ScalaBL (SVD) & $9.656$ & $\underline{0.56}_{\pm 0.0}$ & $\underline{0.32}_{\pm 0.0}$ & $\underline{0.05}_{\pm 0.0}$ & $\underline{0.28}_{\pm 0.0}$ & $\underline{0.12}_{\pm 0.0}$ & $\underline{0.18}_{\pm 0.0}$ \\
& ScalaBL (ours) & $9.648$ & $\underline{0.65}_{\pm 0.0}$ & $\underline{0.32}_{\pm 0.0}$ & $\underline{0.05}_{\pm 0.0}$ & $\underline{0.30}_{\pm 0.0}$ & $\underline{0.12}_{\pm 0.0}$ & $\underline{0.18}_{\pm 0.0}$ \\
\bottomrule
\end{tabular}
\label{tab:qwen32B_main}
\end{table*}

%% file: conclusion.tex
\section{Limitations}\label{sec:limitations}
Regardless of the parameter efficiency of ScalaBL, computing the Bayesian model average over the projected weight samples has the same runtime cost as BLoB. A limitation that this work shares with \cite{lap} and \cite{blob} is that we only consider multiple choice classification datasets for evaluation. This underscores the need for uncertainty quantification of open-ended LLM generations in future work. 

\section{Conclusion}\label{sec:conclusions}
In this work we presented \textbf{Scala}ble \textbf{B}ayesian \textbf{L}ow Rank Adaptation  via Stochastic Variational Subspace Inference (ScalaBL). We perform Bayesian inference over an $r$-dimensional subspace and repurpose the $\A$ and $\B$ parameters of a LoRA as projection matrices which map samples from this low dimensional subspace into the full weight space $\W$ of an LLM. We showed how we can learn all the parameters of our approach using stochastic variational inference. Due to the small dimensionality of our subspace, we enjoy considerable parameter efficiency compared to prior work, while still achieving competitive performance with state-of-the-art approaches on a variety of commonsense reasoning benchmarks. For this reason 
our work is the first to scale a Bayesian LoRA approach to a 32 billion model, while requiring several orders of magnitude fewer additional parameters than prior work.

%% file: appendix.tex
\section{Additional Implementation Details}
\subsection{Uncertainty Quantification Metrics}\label{sec:metric}
In our experiments we report results on two popular metrics for measuring uncertainty quantification: negative log likelihood (NLL) and expected calibration error (ECE).  For dataset of $N$ test instances $\x_n$ with correct label $y_n$ and a probabilistic model $P_{\btheta}$, NLL is defined as:
\begin{align}
    \text{NLL}=\frac{1}{N} \sum_{n=1}^N - \log P_{\btheta} (y_n|\x_n)
\end{align}
That is, it is the expected negative log probability of the correct class under the model.

ECE measures how a model's confidence aligns with the accuracy of its predictions. It can be computed by binning the predictions by their confidence. We then compute a weighted average of the difference between the accuracy and confidence within each bin:
\begin{align}
\text{ECE} = \sum_{k=1}^{K} \frac{|B_k|}{N} \left| \text{acc}(B_k) - \text{conf}(B_k) \right|
\end{align}
where $K$ is the number of bins and $B_k$ is the set of samples in the $k$-th bin. Following \cite{blob}, we use $K=15$ in all experiments.

\input{floats/tables/prompts}
\subsection{Prompts}
Following \cite{blob}, the prompts used for each dataset are displayed in Table \ref{tab:prompts}.

\subsection{Runtime Analysis}
\input{floats/tables/runtime}
The main efficiency savings gained by ScalaBL over BLoB is the reduction of the number of parameters that need to be learned. This also translates to gains in performance during training. In Table \ref{tab:runtime} we show the training resource usage for both ScalaBL and BLoB. We see that ScalaBL has lower peak memory usage and trains slightly faster than BLoB. 

\section{Additional Experimental Results}
\subsection{Effect of Number of Samples}\label{sec:samples}
An important hyperparameter for any variational approach is the number of weight samples $N$ to draw when computing the test time Bayesian model average. Using models fine-tuned on the Winogrande-Small dataset we explore different choices for this hyperparameter for both ScalaBL and BLoB. This is shown for all 3 metrics in Figure \ref{fig:combined_results} (Top). We see that performance across all metrics is saturated around $N=10$, validating the choice of \cite{blob}.

Its important to remember that the number of parameters sampled from the variational distribution is much smaller in ScalaBL as compared to BLoB. In Figure \ref{fig:combined_results} (Bottom) we use a log scale plot to compare how many parameters each method has to draw as we increase the number of samples that are performed.
\input{floats/effect_of_samples}

\input{floats/tables/choice_of_subspace}
\subsection{Choice of Subspace}
In this section we consider different choices for the subspace used in method. In Table \ref{tab:qwen7B_subspace}, we present results using the SVD subspace defined in Equation \ref{eq:svdsubspace}. We also include results for an experiment where the $\A$ matrix is frozen during fine-tuning. This is similar to the random subspace approach put forward by \cite{izmailov2020subspace}. 

We first notice that difference in performance between the SVD subspace and the subspace used in ScalaBL is negligible. This isn't surprising as adding the extra $\U$ matrix of parameters does not change the expressive power of the model as discussed in the main paper. An interesting upside of using the random subspace is that it further reduces the number of parameters that need to be learned. We see that for some datasets performance is comparable the subspaces with more parameters. However, on some datasets (such as Winogrande-Medium) there is a considerable reduction in classification accuracy when using a random subspace.

\input{floats/tables/choice_of_cov}
\subsection{Using a Full Rank Covariance}
Following the  prior work, we only considered using a diagonal covariance for $q_{\btheta}(\s)$ in our experiments in the main paper. For the approaches of \cite{lap} and \cite{blob} this is a necessary limitation as instantiating a full rank covariance with millions of dimensions would be intractable. However, the Gaussian distribution used in ScalaBL is only $r$-dimensional. This makes it straightforward to consider using a full rank Gaussian by adding a few more parameters.

We parameterize a full rank covariance matrix $\boldsymbol{\Sigma} $ as an eigen decomposition. We treat the eigenvalues $\mathbf{e} \in \R^{r}$ and  matrix of eigenvectors $\mathbf{E} \in \R^{r \times r}$ as learnable parameters. This adds $r + r^2$ additional parameters to the approach. We use the QR factorization to ensure that eigenvalues are orthogonal. 
\begin{align}
    \mathbf{E}, \mathbf{R} = \text{QR}(\mathbf{\hat{E}})\\
    \boldsymbol{\Sigma} = \mathbf{E} \diag{\mathbf{e}} \mathbf{E}^T
\end{align}
where $\mathbf{\hat{E}}$ are free parameters.

We then update the reparameterization trick to use the Cholesky factor of the covariance matrix. We apply the Cholesky factorization on-the-fly during learning. 
\begin{align}
    \mathbf{L} = \text{Cholesky}(\boldsymbol{\Sigma} )\\
    \W_t = \W_0 + \B \diag{\s_{\mu} + \mathbf{L}\eps_t}\A
\end{align}

We compare using a diagonal and a full rank covariance matrix in Table \ref{tab:qwen7B_cov}. We see that using a full rank covariance leads to very similar performance to using a diagonal one, with some datasets even exhibiting worse calibration.

\subsection{Effect of LoRA Rank}
In the prior work of \cite{lap} and \cite{blob} a LoRA rank of $r=8$ was used in all experiments. In the main paper, we use this value for the rank as well. In this section we explore the effect of the LoRA on performance for both ScalaBL and BLoB. We ran additional in-distribution experiments using Qwen2.5-7B with $r = [4,16,32]$ to compare against the results for $r=8$ which are already shown in Table \ref{tab:qwen7B_main}. This results are shown in Figures \ref{fig:r_sweep1} and \ref{fig:r_sweep2}.

We first note that the $x$-axes of these plots show the number of total model parameters, rather than the rank. This captures the fact that BLoB’s additional parameters grow more quickly as $r$ increases ($O(rd$)) compared to ScalaBL ($O(r)$). We see that BLoB often results in noticeable drops in accuracy as $r$ increases across multiple datasets (WG-S, ARC-E, ARC-C, WG-M, OBQA). This is then accompanied by increases in NLL. By comparison ScalaBL sees small increases in accuracy across most datasets as $r$ increases, albeit accompanied with small increases in ECE. Furthermore, BLoB sees larger increases in ECE compared to ScalaBL across multiple datasets (ARC-E, ARC-C, OBQA). The only time that BLoB sees a reduction in ECE is when the accuracy also decreases significantly (WG-S, WG-M). ScalaBL is robust to changes in $r$ across all 3 metrics. 
\input{floats/tables/llama7B_main}
\input{floats/tables/llama7B_ood}

\subsection{Llama2 Results} \label{sec:llama2}
For the sake of comparison with \cite{lap} and \cite{blob}, we present experimental results using the older \texttt{Llama-2-7b} LLM in Tables \ref{tab:llama2_main} and \ref{tab:llama2_ood}. We note that we reran BLoB and the standard baselines using 16-bit frozen parameters, instead of 8-bit quantized weights. The reported results for Laplace and Bayes By Backprop (BBB) \citep{bbb} are repeated from the tables of \cite{blob}. We obverse the same general trends as seen with \texttt{Qwen2.5-7B}. Our proposed approach achieves competitive performance with BLoB while requiring significantly fewer parameters.
\input{floats/r_sweep}

%% file: floats/tables/prompts.tex
\begin{table} 
\centering
\caption{Prompts used for each dataset.}
\begin{tabular}{c|c}
\toprule
\textbf{Dataset(s)}            & \textbf{Prompt }  \\
\midrule 
\textbf{WG-S}, \textbf{WG-M}            & Select one of the choices that answer the following question: \{question\}  \\
& Choices: A. \{option1\}. B. \{option2\}. Answer:                              \\
\midrule
\textbf{ARC-E},\textbf{ARC-C}& Select one of the choices that answer the following question: \{question\} \\
\textbf{OBQA},\textbf{MMLU}  & Choices: A. \{option1\}. B. \{option2\}. C. \{option3\}. D. \{option4\}. Answer:   \\
\midrule
\textbf{BoolQ }                & Answer the question with only True or False: \{question\} Context: \{passage\}.\\
\bottomrule
\end{tabular}
\label{tab:prompts}
\end{table}

%% file: floats/tables/runtime.tex
\begin{table*}
\centering
\begin{tabular}{|c|c|c|c|c|}
\hline
\textbf{Method} & \textbf{Model Size} & \textbf{Batch Size} & \textbf{Peak Memory Usage (GB)} & \textbf{Total Training Time (s)} \\ \hline
ScalaBL         & 7B                  & 4                   & 22.79                           & 1050                             \\
BLoB            & 7B                  & 4                   & 23.47                           & 1064                             \\
ScalaBL         & 32B                 & 2                   & 72.36                           & 2560                             \\
BLoB            & 32B                 & 2                   & 74.04                           & 2740          \\
\hline
\end{tabular}
\caption{Training-time resource usage comparison for Winogrande-Small (WG-S) dataset.}
\label{tab:runtime}
\end{table*}

%% file: floats/effect_of_samples.tex
\begin{figure*}[h!]
  \centering
  % ---------- first row: vary # test-time samples ----------
  \begin{minipage}{0.32\textwidth}\centering
    \includegraphics[width=\textwidth]{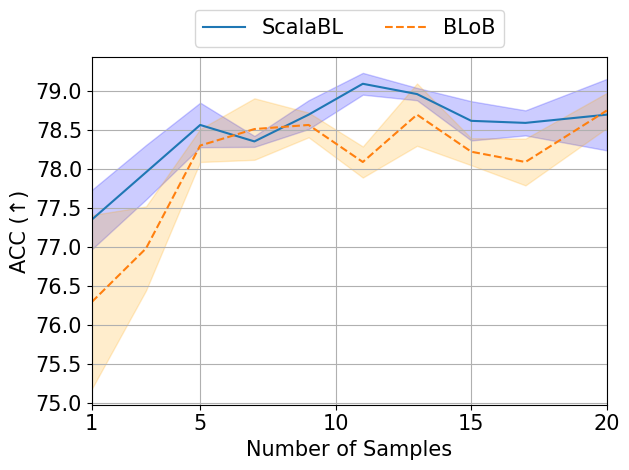}
  \end{minipage}\hfill
  \begin{minipage}{0.32\textwidth}\centering
    \includegraphics[width=\textwidth]{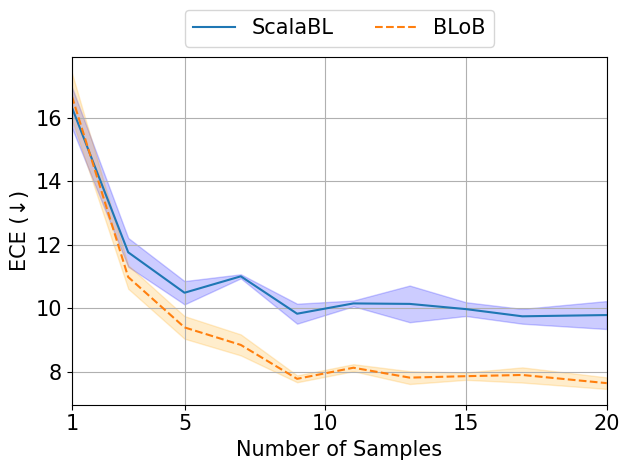}
  \end{minipage}\hfill
  \begin{minipage}{0.32\textwidth}\centering
    \includegraphics[width=\textwidth]{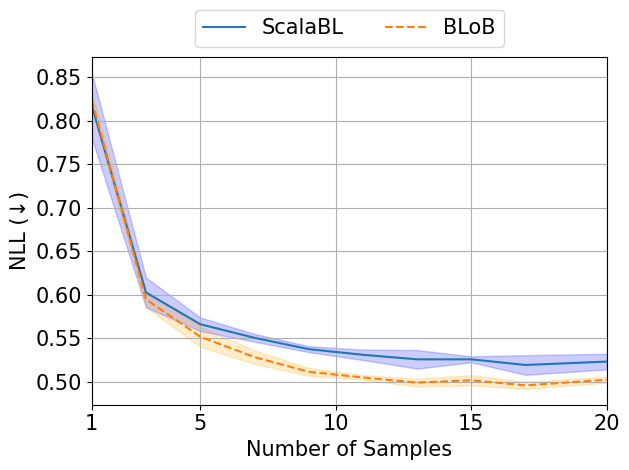}
  \end{minipage}  % small vertical gap

  % ---------- second row: vary total # sampled parameters ----------
  \begin{minipage}{0.32\textwidth}\centering
    \includegraphics[width=\textwidth]{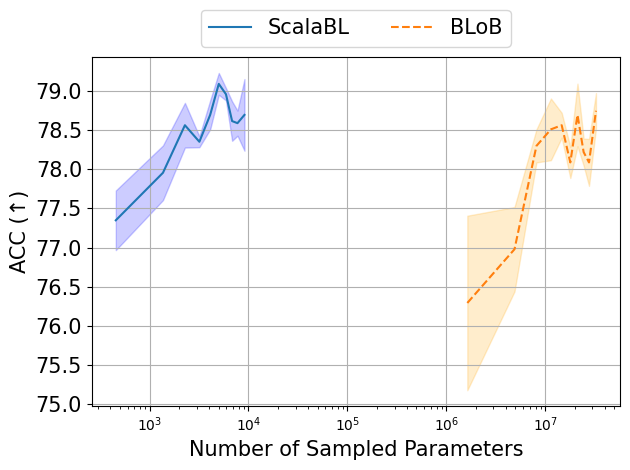}
  \end{minipage}\hfill
  \begin{minipage}{0.32\textwidth}\centering
    \includegraphics[width=\textwidth]{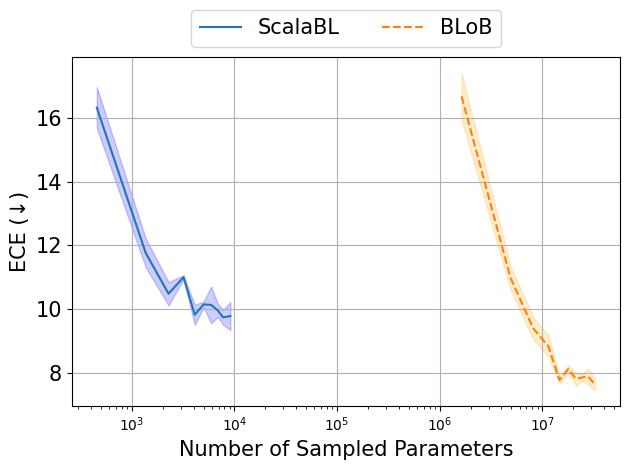}
  \end{minipage}\hfill
  \begin{minipage}{0.32\textwidth}\centering
    \includegraphics[width=\textwidth]{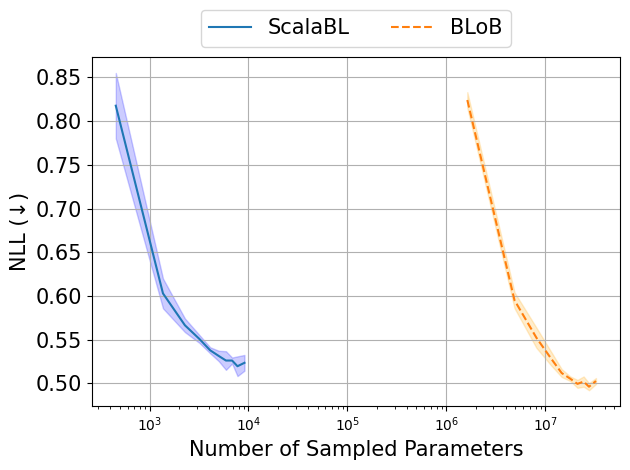}
  \end{minipage}

  \caption{Effect of number of variational samples on Winogrande-Small (WG-S). 
  \textbf{Top row:} number of test-time samples. 
  \textbf{Bottom row:} total number of sampled parameters. 
  Shaded areas show the standard error over three training seeds.}
  \label{fig:combined_results}
\end{figure*}

%% file: floats/tables/choice_of_subspace.tex
\begin{table*}
\centering
\caption{
In-distribution experiment using \texttt{Qwen2.5-7B} using difference choices for the subpace.
We report the mean and standard deviation of test set performance using 3 training seeds.
}
\begin{tabular}{@{}ccc|cccccc@{}}
\toprule
\multirow{2}{*}{\textbf{Metric}} & \multirow{2}{*}{\textbf{Subspace}}  & \multirow{2}{*}{\textbf{Params (M)}} & \multicolumn{6}{c}{\textbf{Datasets}} \\ 
  &  & & \textbf{WG-S} & \textbf{ARC-C} & \textbf{ARC-E} & \textbf{WG-M} & \textbf{OBQA} & \textbf{BoolQ} \\
\midrule
\multirow{3}{*}{\textbf{ACC ($\uparrow$)}}
& Random & $2.135$ & $73.18_{\pm 0.4}$ & $89.30_{\pm 0.2}$ & $95.66_{\pm 0.4}$ & $74.60_{\pm 0.4}$ & $86.80_{\pm 0.6}$ & $86.67_{\pm 0.3}$ \\
& SVD & $3.773$ & $78.90_{\pm 0.4}$ & $89.30_{\pm 0.5}$ & $96.30_{\pm 0.5}$ & $81.62_{\pm 0.4}$ & $91.07_{\pm 0.5}$ & $88.59_{\pm 0.1}$ \\
& ScalaBL & $3.769$ & $78.64_{\pm 0.4}$ & $90.16_{\pm 0.8}$ & $96.26_{\pm 0.1}$ & $81.42_{\pm 0.3}$ & $90.90_{\pm 0.5}$ & $88.48_{\pm 0.1}$ \\
\midrule
\multirow{3}{*}{\textbf{ECE ($\downarrow$)}}
& Random & $2.135$ & $12.96_{\pm 0.2}$ & $4.75_{\pm 0.7}$ & $2.08_{\pm 0.2}$ & $6.72_{\pm 0.8}$ & $2.89_{\pm 0.4}$ & $1.30_{\pm 0.2}$ \\
& SVD & $3.773$ & $9.93_{\pm 0.2}$ & $6.01_{\pm 0.5}$ & $2.34_{\pm 0.3}$ & $5.43_{\pm 0.4}$ & $2.57_{\pm 0.6}$ & $1.31_{\pm 0.1}$\\ 
& ScalaBL & $3.769$ & $8.88_{\pm 0.5}$ & $5.03_{\pm 0.9}$ & $1.78_{\pm 0.2}$ & $3.64_{\pm 0.2}$ & $2.43_{\pm 0.7}$ & $1.96_{\pm 0.3}$
\\
\midrule
\multirow{3}{*}{\textbf{NLL ($\downarrow$)}}
& Random & $2.135$ & $0.62_{\pm 0.0}$ & $0.31_{\pm 0.0}$ & $0.12_{\pm 0.0}$ & $0.53_{\pm 0.0}$ & $0.37_{\pm 0.0}$ & $0.27_{\pm 0.0}$ \\
& SVD & $3.773$ & $0.53_{\pm 0.0}$ & $0.35_{\pm 0.0}$ & $0.12_{\pm 0.0}$ & $0.39_{\pm 0.0}$ & $0.22_{\pm 0.0}$ & $0.23_{\pm 0.0}$ \\
& ScalaBL & $3.769$ & $0.51_{\pm 0.0}$ & $0.31_{\pm 0.0}$ & $0.11_{\pm 0.0}$ & $0.40_{\pm 0.0}$ & $0.23_{\pm 0.0}$ & $0.24_{\pm 0.0}$ \\
\bottomrule
\end{tabular}
\label{tab:qwen7B_subspace}
\end{table*}

%% file: floats/tables/choice_of_cov.tex
\begin{table*}
\centering
\caption{
In-distribution experiment using \texttt{Qwen2.5-7B} using difference choices for the covariance of $q_{\btheta}(\s)$.
We report the mean and standard deviation of test set performance using 3 training seeds.
}
\begin{tabular}{@{}ccc|cccccc@{}}
\toprule
\multirow{2}{*}{\textbf{Metric}} & \multirow{2}{*}{$\boldsymbol{\Sigma}$}  & \multirow{2}{*}{\textbf{Params (M)}} & \multicolumn{6}{c}{\textbf{Datasets}} \\ 
  &  & & \textbf{WG-S} & \textbf{ARC-C} & \textbf{ARC-E} & \textbf{WG-M} & \textbf{OBQA} & \textbf{BoolQ} \\
\midrule
\multirow{2}{*}{\textbf{ACC ($\uparrow$)}}
& Full Rank & $3.773$ & $77.93_{\pm 0.3}$ & $89.30_{\pm 0.5}$ & $96.48_{\pm 0.2}$ & $81.88_{\pm 0.5}$ & $91.73_{\pm 0.6}$ & $88.74_{\pm 0.1}$ \\
& Diagonal & $3.769$ & $78.64_{\pm 0.4}$ & $90.16_{\pm 0.8}$ & $96.26_{\pm 0.1}$ & $81.42_{\pm 0.3}$ & $90.90_{\pm 0.5}$ & $88.48_{\pm 0.1}$ \\
\midrule
\multirow{2}{*}{\textbf{ECE ($\downarrow$)}}
& Full Rank & $3.773$ & $13.25_{\pm 0.6}$ & $6.69_{\pm 0.6}$ & $2.65_{\pm 0.1}$ & $7.11_{\pm 0.2}$ & $2.61_{\pm 0.2}$ & $1.57_{\pm 0.2}$ \\
& Diagonal & $3.769$ & $8.88_{\pm 0.5}$ & $5.03_{\pm 0.9}$ & $1.78_{\pm 0.2}$ & $3.64_{\pm 0.2}$ & $2.43_{\pm 0.7}$ & $1.96_{\pm 0.3}$
\\
\midrule
\multirow{2}{*}{\textbf{NLL ($\downarrow$)}}
& Full Rank & $3.773$ & $0.67_{\pm 0.0}$ & $0.41_{\pm 0.0}$ & $0.14_{\pm 0.0}$ & $0.42_{\pm 0.0}$ & $0.23_{\pm 0.0}$ & $0.23_{\pm 0.0}$ \\
& Diagonal & $3.769$ & $0.51_{\pm 0.0}$ & $0.31_{\pm 0.0}$ & $0.11_{\pm 0.0}$ & $0.40_{\pm 0.0}$ & $0.23_{\pm 0.0}$ & $0.24_{\pm 0.0}$ \\

\bottomrule
\end{tabular}
\label{tab:qwen7B_cov}
\end{table*}

%% file: floats/tables/llama7B_main.tex
\begin{table*}
\centering
\caption{In-distribution experiment using \texttt{Llama-2-7b}. 
We report the mean and standard deviation of test set performance using 3 training seeds.
\textbf{Bold} and \underline{underlined} results denote the best and second best mean performance on each metric/dataset.
}
\begin{tabular}{@{}ccc|cccccc@{}}
\toprule
\multirow{2}{*}{\textbf{Metric}} & \multirow{2}{*}{\textbf{Method}}  & \multirow{2}{*}{\textbf{Params (M)}} & \multicolumn{6}{c}{\textbf{Datasets}} \\ 
  &  & & \textbf{WG-S} & \textbf{ARC-C} & \textbf{ARC-E} & \textbf{WG-M} & \textbf{OBQA} & \textbf{BoolQ} \\
\midrule
\multirow{8}{*}{\textbf{ACC ($\uparrow$)}}
& MLE & $4.483$ & $70.78_{\pm 1.2}$ & $\underline{70.16}_{\pm 2.8}$ & $86.97_{\pm 0.6}$ & $75.26_{\pm 0.6}$ & $82.53_{\pm 0.4}$ & $88.02_{\pm 0.2}$ \\
& MAP & $4.483$ & $\underline{71.10}_{\pm 1.1}$ & $68.81_{\pm 0.2}$ & $86.62_{\pm 0.5}$ & $\textbf{76.45}_{\pm 0.7}$ & $82.80_{\pm 0.2}$ & $\underline{88.05}_{\pm 0.2}$ \\
& MC-Dropout & $4.483$ & $69.99_{\pm 2.7}$ & $68.13_{\pm 1.0}$ & $\underline{87.68}_{\pm 0.4}$ & $76.05_{\pm 0.7}$ & $\underline{83.07}_{\pm 1.2}$ & $\textbf{88.43}_{\pm 0.3}$ \\
& Ensemble & $13.449$ & $\textbf{71.31}_{\pm 0.3}$ & $\textbf{71.17}_{\pm 1.4}$ & $\textbf{88.32}_{\pm 0.3}$ & $\underline{76.37}_{\pm 0.8}$ & $\textbf{83.53}_{\pm 0.2}$ & $87.87_{\pm 0.2}$ \\
& BBB & $6.613$ & $56.54_{\pm 0.7}$ & $68.13_{\pm 1.3}$ & $85.86_{\pm 0.7}$ & $73.63_{\pm 2.4}$ & $82.06_{\pm 0.6}$ & $87.21_{\pm 0.2}$ \\
& Laplace & $4.483$ & $69.20_{\pm 1.5}$ & $66.78_{\pm 0.7}$ & $80.05_{\pm 0.2}$ & $75.55_{\pm 0.4}$ & $82.12_{\pm 0.7}$ & $86.95_{\pm 0.1}$ \\
& BLoB & $6.613$ & $64.93_{\pm 5.1}$ & $70.02_{\pm 0.9}$ & $85.80_{\pm 0.6}$ & $73.71_{\pm 1.4}$ & $82.47_{\pm 0.4}$ & $87.62_{\pm 0.2}$ \\
%& ScalaBL (SVD)  & $4.488$ & $70.15_{\pm 1.3}$ & $68.02_{\pm 2.2}$ & $86.27_{\pm 0.6}$ & $74.92_{\pm 1.1}$ & $81.60_{\pm 0.9}$ & $86.75_{\pm 0.2}$ \\
& ScalaBL (ours) & $4.488$ & $70.23_{\pm 0.9}$ & $68.58_{\pm 1.8}$ & $86.80_{\pm 0.5}$ & $74.45_{\pm 0.9}$ & $82.13_{\pm 0.2}$ & $86.50_{\pm 0.2}$ \\
\midrule
\multirow{8}{*}{\textbf{ECE ($\downarrow$)}}
& MLE & $4.483$ & $28.17_{\pm 1.5}$ & $28.26_{\pm 1.9}$ & $12.03_{\pm 0.9}$ & $21.97_{\pm 0.6}$ & $13.86_{\pm 0.5}$ & $4.57_{\pm 0.3}$ \\
& MAP & $4.483$ & $27.77_{\pm 1.4}$ & $29.80_{\pm 0.3}$ & $11.82_{\pm 0.1}$ & $21.08_{\pm 0.3}$ & $13.91_{\pm 0.3}$ & $4.34_{\pm 0.2}$ \\
& MC-Dropout & $4.483$ & $28.55_{\pm 2.5}$ & $29.43_{\pm 1.1}$ & $11.25_{\pm 0.3}$ & $20.69_{\pm 0.6}$ & $12.94_{\pm 1.2}$ & $4.17_{\pm 0.1}$ \\
& Ensemble & $13.449$ & $24.61_{\pm 0.5}$ & $25.10_{\pm 1.3}$ & $10.02_{\pm 0.1}$ & $16.96_{\pm 0.6}$ & $10.81_{\pm 0.2}$ & $3.05_{\pm 0.2}$ \\
& BBB & $6.613$ & $21.81_{\pm 13.0}$ & $26.23_{\pm 1.5}$ & $12.28_{\pm 0.6}$ & $15.76_{\pm 4.7}$ & $11.38_{\pm 1.1}$ & $3.74_{\pm 0.1}$ \\
& Laplace & $4.483$ & $\textbf{4.15}_{\pm 1.12}$ & $16.25_{\pm 2.6}$ & $33.29_{\pm 0.6}$ & $7.40_{\pm 0.3}$ & $8.70_{\pm 1.8}$ & $\underline{1.30}_{\pm 0.3}$ \\
& BLoB & $6.613$ & $\underline{5.55}_{\pm 3.3}$ & $\underline{14.05}_{\pm 0.7}$ & $\underline{3.39}_{\pm 1.0}$ & $\textbf{3.36}_{\pm 0.5}$ & $\textbf{2.80}_{\pm 0.5}$ & $\textbf{1.08}_{\pm 0.2}$ \\
%& ScalaBL (SVD)  & $4.488$ & $9.24_{\pm 1.2}$ & $\textbf{11.78}_{\pm 1.2}$ & $\textbf{3.25}_{\pm 0.2}$ & $\underline{3.47}_{\pm 0.3}$ & $\underline{4.02}_{\pm 0.4}$ & $2.07_{\pm 0.2}$ \\
& ScalaBL (ours) & $4.488$ & $9.49_{\pm 1.2}$ & $\textbf{9.79}_{\pm 1.9}$ & $\textbf{3.54}_{\pm 0.2}$ & $\underline{4.31}_{\pm 0.4}$ & $\underline{3.62}_{\pm 0.9}$ & $1.83_{\pm 0.3}$ \\
\midrule
\multirow{8}{*}{\textbf{NLL ($\downarrow$)}}
& MLE & $4.483$ & $2.47_{\pm 0.2}$ & $2.71_{\pm 0.4}$ & $0.98_{\pm 0.1}$ & $1.14_{\pm 0.1}$ & $0.91_{\pm 0.1}$ & $0.27_{\pm 0.0}$ \\
& MAP & $4.483$ & $2.89_{\pm 0.5}$ & $3.02_{\pm 0.2}$ & $1.05_{\pm 0.0}$ & $1.14_{\pm 0.0}$ & $0.89_{\pm 0.0}$ & $0.27_{\pm 0.0}$ \\
& MC-Dropout & $4.483$ & $3.01_{\pm 0.4}$ & $3.08_{\pm 0.1}$ & $1.00_{\pm 0.1}$ & $1.03_{\pm 0.0}$ & $0.86_{\pm 0.1}$ & $0.27_{\pm 0.0}$ \\
& Ensemble & $13.449$ & $1.47_{\pm 0.0}$ & $1.93_{\pm 0.1}$ & $0.74_{\pm 0.0}$ & $0.73_{\pm 0.0}$ & $0.64_{\pm 0.0}$ & $\underline{0.25}_{\pm 0.0}$ \\
& BBB & $6.613$ & $1.40_{\pm 0.6}$ & $2.23_{\pm 0.0}$ & $0.91_{\pm 0.0}$ & $0.84_{\pm 0.2}$ & $0.66_{\pm 0.1}$ & $0.31_{\pm 0.0}$ \\
& Laplace & $4.483$ & $\textbf{0.60}_{\pm 0.0}$ & $1.03_{\pm 0.0}$ & $0.88_{\pm 0.0}$ & $
\underline{0.57}_{\pm 0.0}$ & $0.52_{\pm 0.0}$ & $0.31_{\pm 0.0}$ \\
& BLoB & $6.613$ & $0.66_{\pm 0.1}$ & $\underline{0.87}_{\pm 0.0}$ & $\textbf{0.38}_{\pm 0.0}$ & $\textbf{0.51}_{\pm 0.0}$ & $\textbf{0.47}_{\pm 0.0}$ & $\textbf{0.23}_{\pm 0.0}$ \\
%& ScalaBL (SVD) & $4.488$ & $\textbf{0.60}_{\pm 0.0}$ & $\textbf{0.79}_{\pm 0.0}$ & $\textbf{0.38}_{\pm 0.0}$ & $\textbf{0.50}_{\pm 0.0}$ & $\underline{0.51}_{\pm 0.0}$ & $\underline{0.25}_{\pm 0.0}$ \\
& ScalaBL (ours) & $4.488$ & $\underline{0.59}_{\pm 0.0}$ & $\underline{0.79}_{\pm 0.0}$ & $\underline{0.39}_{\pm 0.0}$ & $\textbf{0.51}_{\pm 0.0}$ & $\underline{0.51}_{\pm 0.0}$ & $\underline{0.25}_{\pm 0.0}$ \\
\bottomrule
\end{tabular}
\label{tab:llama2_main}
\end{table*}

%% file: floats/tables/llama7B_ood.tex
\begin{table*}
\centering
\caption{Out-of-distribution experiment using \texttt{Llama-2-7b}. 
We report the mean and standard deviation of test set performance using 3 training seeds.
\textbf{Bold} and \underline{underlined} results denote the best and second best mean performance on each metric/dataset.}
\begin{tabular}{@{}ccc|c|cc|ccc@{}}
\toprule
\multirow{3}{*}{\textbf{Metric}} & \multirow{3}{*}{\textbf{Method}}  & \multirow{3}{*}{\textbf{Params (M)}} & \multicolumn{5}{c}{\textbf{Datasets}} \\ 

& & &  \multicolumn{1}{c}{\textbf{In Dist.}} & \multicolumn{2}{c}{\textbf{Smaller Dist. Shift}} & \multicolumn{2}{c}{\textbf{Larger Dist. Shift}} \\ 
  \cline{4-9}
  &  & & \textbf{OBQA} & \textbf{ARC-C} & \textbf{ARC-E} & \textbf{Chemistry} & \textbf{Physics} & \\
\midrule
\multirow{8}{*}{\textbf{ACC ($\uparrow$)}}
& MLE & $4.483$ & $82.53_{\pm 0.4}$ & $\underline{69.48}_{\pm 0.5}$ & $75.59_{\pm 1.2}$ & $39.33_{\pm 1.5}$ & $29.00_{\pm 2.6}$ \\
& MAP & $4.483$ & $\underline{82.80}_{\pm 0.2}$ & $68.92_{\pm 1.2}$ & $76.29_{\pm 0.7}$ & $36.00_{\pm 1.0}$ & $\underline{31.00}_{\pm 1.0}$ \\
& MC-Dropout & $4.483$ & $83.07_{\pm 1.2}$ & $69.14_{\pm 0.5}$ & $76.17_{\pm 0.9}$ & $37.67_{\pm 2.1}$ & $28.00_{\pm 4.4}$ \\
& Ensemble & $13.449$ & $\textbf{83.53}_{\pm 0.2}$ & $69.37_{\pm 0.5}$ & $76.12_{\pm 1.0}$ & $38.33_{\pm 1.5}$ & $29.00_{\pm 2.6}$ \\
& BBB & $6.613$ & $82.06_{\pm 0.6}$ & $67.25_{\pm 1.2}$ & $75.83_{\pm 0.8}$ & $42.36_{\pm 0.5}$ & $30.21_{\pm 2.3}$ \\
& Laplace & $4.483$ & $82.12_{\pm 0.7}$ & $69.14_{\pm 1.2}$ & $74.94_{\pm 1.0}$ & $\textbf{44.10}_{\pm 1.3}$ & $\textbf{31.60}_{\pm 0.5}$ \\
& BLoB & $6.613$ & $82.47_{\pm 0.4}$ & $\textbf{69.56}_{\pm 1.1}$ & $\underline{76.55}_{\pm 0.3}$ & $\underline{43.40}_{\pm 0.6}$ & $30.56_{\pm 1.2}$ \\
%& ScalaBL (SVD) & $4.488$ & $81.60_{\pm 0.9}$ & $67.94_{\pm 1.1}$ & $76.13_{\pm 1.4}$ & $\underline{43.40}_{\pm 1.2}$ & $28.82_{\pm 0.6}$ \\
& ScalaBL (ours) & $4.484$ & $82.13_{\pm 0.2}$ & $\underline{69.48}_{\pm 0.5}$ & $\textbf{77.46}_{\pm 0.3}$ & $42.00_{\pm 2.6}$ & $30.33_{\pm 0.6}$ \\
\midrule
\multirow{8}{*}{\textbf{ECE ($\downarrow$)}}
& MLE & $4.483$ & $13.86_{\pm 0.5}$ & $23.07_{\pm 0.9}$ & $17.41_{\pm 0.9}$ & $22.56_{\pm 2.5}$ & $29.36_{\pm 2.3}$ \\
& MAP & $4.483$ & $13.91_{\pm 0.3}$ & $24.10_{\pm 0.9}$ & $16.93_{\pm 1.0}$ & $25.96_{\pm 2.0}$ & $28.30_{\pm 2.5}$ \\
& MC-Dropout & $4.483$ & $12.94_{\pm 1.2}$ & $23.44_{\pm 0.7}$ & $16.84_{\pm 0.7}$ & $23.78_{\pm 3.0}$ & $32.71_{\pm 4.0}$ \\
& Ensemble & $13.449$ & $10.81_{\pm 0.2}$ & $19.12_{\pm 1.1}$ & $13.66_{\pm 0.9}$ & $15.94_{\pm 1.5}$ & $\underline{20.86}_{\pm 2.5}$ \\
& BBB & $6.613$ & $11.38_{\pm 1.1}$ & $19.90_{\pm 0.7}$ & $13.41_{\pm 0.9}$ & $15.67_{\pm 1.2}$ & $26.10_{\pm 4.8}$ \\
& Laplace & $4.483$ & $8.70_{\pm 1.8}$ & $\textbf{5.84}_{\pm 0.6}$ & $\underline{8.51}_{\pm 1.1}$ & $\textbf{10.76}_{\pm 3.4}$ & $\textbf{13.91}_{\pm 0.9}$ \\
& BLoB & $6.613$ & $\textbf{2.80}_{\pm 0.5}$ & $13.82_{\pm 0.5}$ & $9.65_{\pm 0.7}$ & $\underline{15.39}_{\pm 3.4}$ & $22.66_{\pm 0.7}$ \\
%$& ScalaBL (SVD) & $4.488$ & $\underline{4.02}_{\pm 0.4}$ & $\underline{11.69}_{\pm 0.8}$ & $\textbf{7.96}_{\pm 1.1}$ & $\underline{15.31}_{\pm 1.1}$ & $23.90_{\pm 0.8}$ \\
& ScalaBL (ours) & $4.484$ & $\underline{3.62}_{\pm 0.9}$ & $\underline{11.85}_{\pm 0.6}$ & $\textbf{7.89}_{\pm 0.8}$ & $15.99_{\pm 3.3}$ & $21.98_{\pm 1.1}$ \\
\midrule
\multirow{8}{*}{\textbf{NLL ($\downarrow$)}}
& MLE & $4.483$ & $0.91_{\pm 0.1}$ & $1.42_{\pm 0.1}$ & $1.11_{\pm 0.1}$ & $1.62_{\pm 0.0}$ & $1.69_{\pm 0.1}$ \\
& MAP & $4.483$ & $0.89_{\pm 0.0}$ & $1.46_{\pm 0.1}$ & $1.12_{\pm 0.0}$ & $1.67_{\pm 0.1}$ & $1.70_{\pm 0.1}$ \\
& MC-Dropout & $4.483$ & $0.86_{\pm 0.1}$ & $1.39_{\pm 0.1}$ & $1.12_{\pm 0.1}$ & $1.64_{\pm 0.1}$ & $1.76_{\pm 0.0}$ \\
& Ensemble & $13.449$ & $0.64_{\pm 0.0}$ & $1.03_{\pm 0.0}$ & $0.82_{\pm 0.0}$ & $1.42_{\pm 0.0}$ & $1.49_{\pm 0.0}$ \\
& BBB & $6.613$ & $0.66_{\pm 0.1}$ & $1.06_{\pm 0.0}$ & $0.79_{\pm 0.0}$ & $1.49_{\pm 0.0}$ & $1.62_{\pm 0.1}$ \\
& Laplace & $4.483$ & $0.52_{\pm 0.0}$ & $\textbf{0.81}_{\pm 0.0}$ & $\underline{0.70}_{\pm 0.0}$ & $\textbf{1.35}_{\pm 0.0}$ & $\textbf{1.36}_{\pm 0.0}$ \\
& BLoB & $6.613$ & $\textbf{0.47}_{\pm 0.0}$ & $0.88_{\pm 0.0}$ & $\underline{0.70}_{\pm 0.0}$ & $\underline{1.38}_{\pm 0.0}$ & $\underline{1.43}_{\pm 0.0}$ \\
%& ScalaBL (SVD)& $4.488$ & $\underline{0.51}_{\pm 0.0}$ & $\underline{0.88}_{\pm 0.0}$ & $\textbf{0.65}_{\pm 0.0}$ & $1.39_{\pm 0.0}$ & $1.50_{\pm 0.0}$ \\
& ScalaBL (ours) & $4.484$ & $\underline{0.51}_{\pm 0.0}$ & $\underline{0.85}_{\pm 0.0}$ & $\textbf{0.63}_{\pm 0.0}$ & $1.40_{\pm 0.0}$ & $1.48_{\pm 0.0}$ \\
\bottomrule
\end{tabular}
\label{tab:llama2_ood}
\end{table*}

%% file: floats/r_sweep.tex
\begin{figure*}[h!]
  \centering
  % ---------- Winogrande-Small ----------
  \begin{minipage}{0.32\textwidth}\centering
    \includegraphics[width=\textwidth]{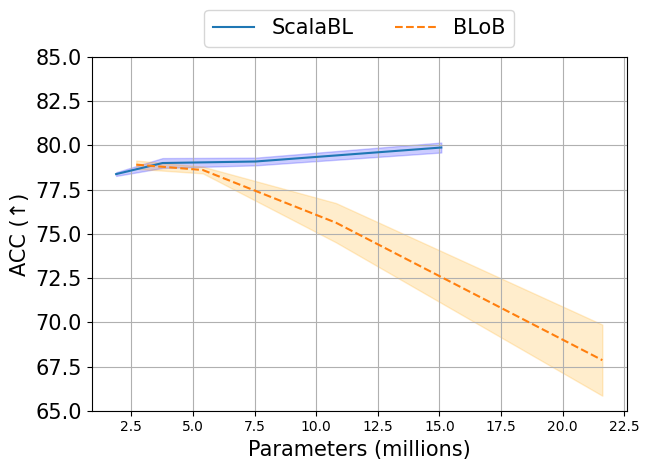}
  \end{minipage}\hfill
  \begin{minipage}{0.32\textwidth}\centering
    \includegraphics[width=\textwidth]{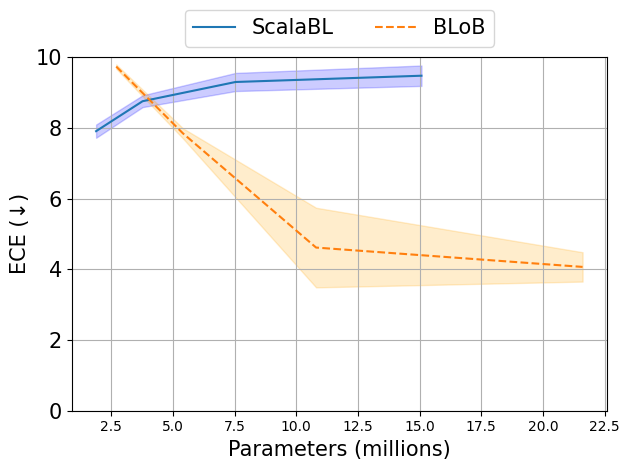}
  \end{minipage}\hfill
  \begin{minipage}{0.32\textwidth}\centering
    \includegraphics[width=\textwidth]{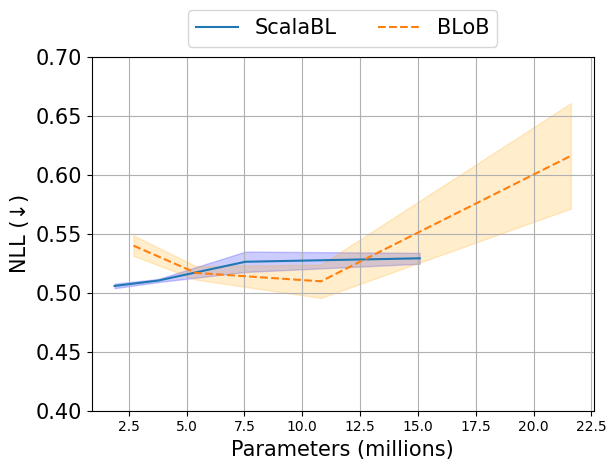}
  \end{minipage}\\[-0.5em]

  % ---------- ARC-Easy ----------
  \begin{minipage}{0.32\textwidth}\centering
    \includegraphics[width=\textwidth]{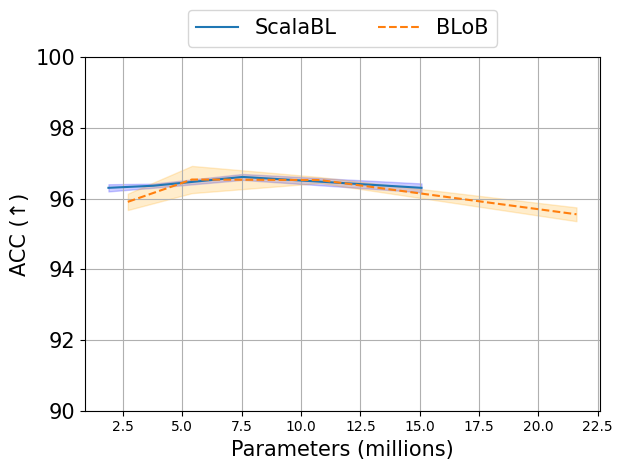}
  \end{minipage}\hfill
  \begin{minipage}{0.32\textwidth}\centering
    \includegraphics[width=\textwidth]{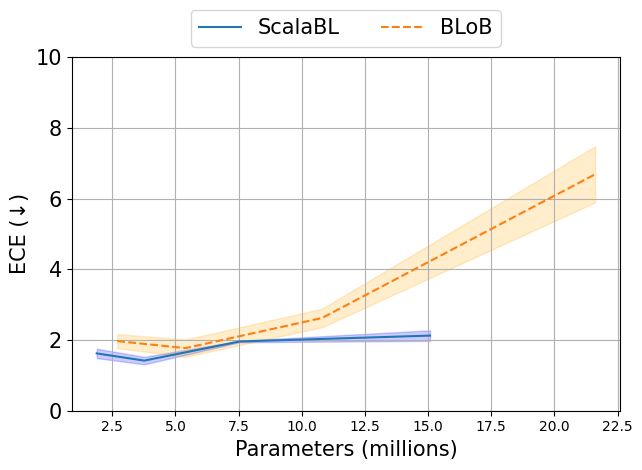}
  \end{minipage}\hfill
  \begin{minipage}{0.32\textwidth}\centering
    \includegraphics[width=\textwidth]{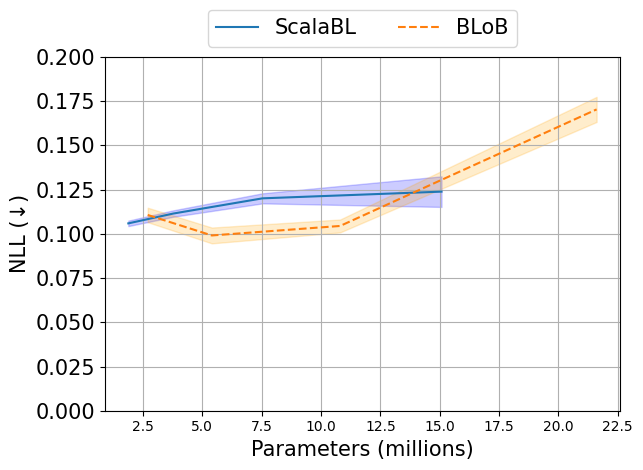}
  \end{minipage}\\[-0.5em]

  % ---------- ARC-Challenge ----------
  \begin{minipage}{0.32\textwidth}\centering
    \includegraphics[width=\textwidth]{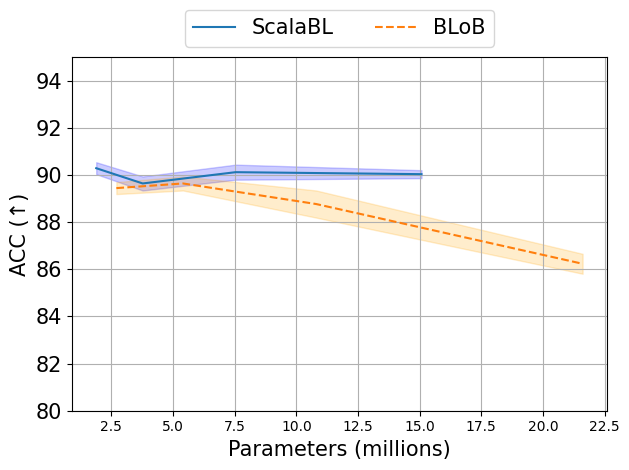}
  \end{minipage}\hfill
  \begin{minipage}{0.32\textwidth}\centering
    \includegraphics[width=\textwidth]{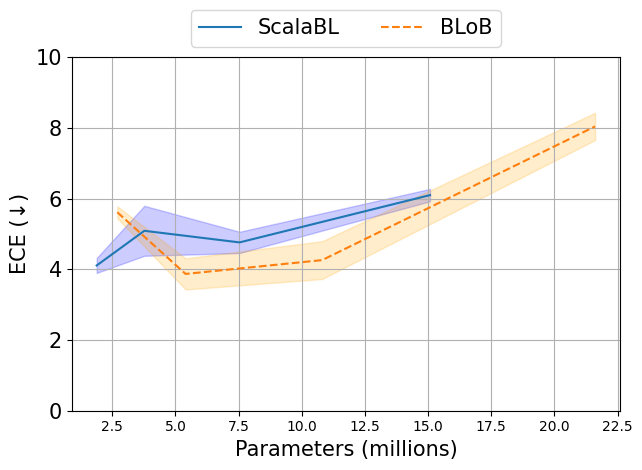}
  \end{minipage}\hfill
  \begin{minipage}{0.32\textwidth}\centering
    \includegraphics[width=\textwidth]{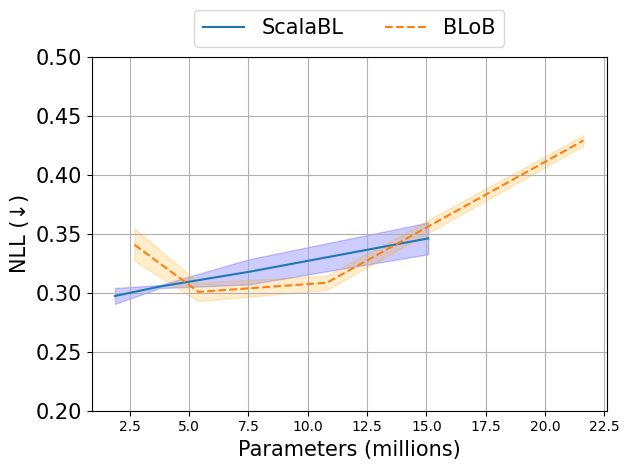}
  \end{minipage}

  \caption{Effect of LoRA rank $r$. \textbf{Top:} Winogrande-Small (WG-S). \textbf{Middle:} ARC-Easy (ARC-E). \textbf{Bottom:} ARC-Challenge (ARC-C).}
  \label{fig:r_sweep1}
\end{figure*}

\begin{figure*}[h!]
  \centering
  % ---------- Winogrande-Medium ----------
  \begin{minipage}{0.32\textwidth}\centering
    \includegraphics[width=\textwidth]{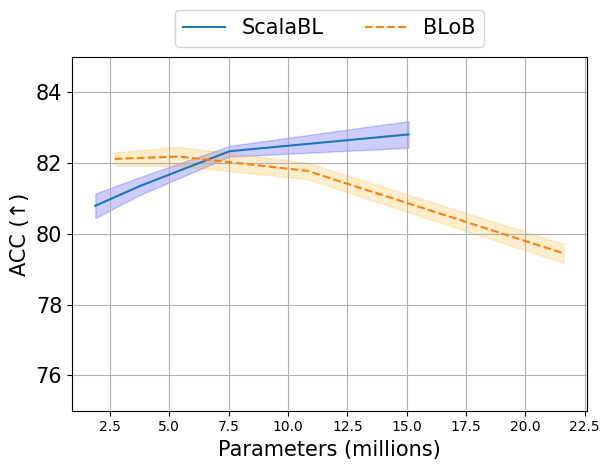}
  \end{minipage}\hfill
  \begin{minipage}{0.32\textwidth}\centering
    \includegraphics[width=\textwidth]{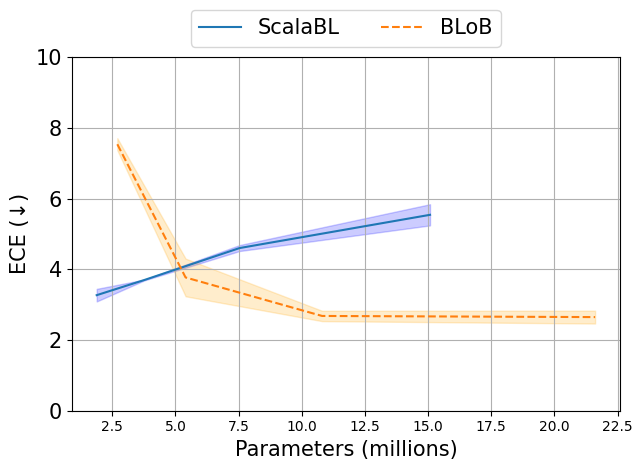}
  \end{minipage}\hfill
  \begin{minipage}{0.32\textwidth}\centering
    \includegraphics[width=\textwidth]{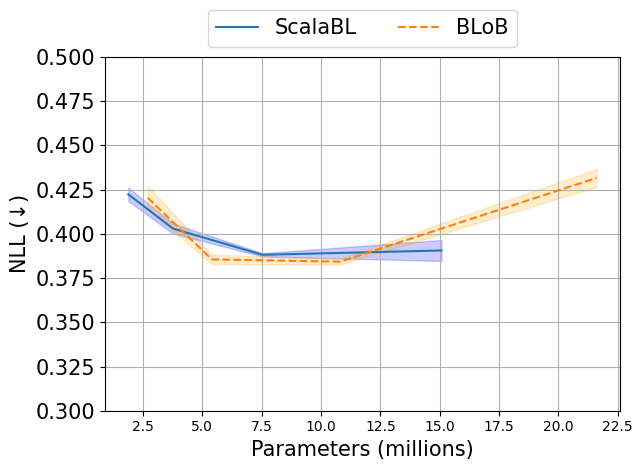}
  \end{minipage}\\[-0.5em]

  % ---------- OpenBookQA ----------
  \begin{minipage}{0.32\textwidth}\centering
    \includegraphics[width=\textwidth]{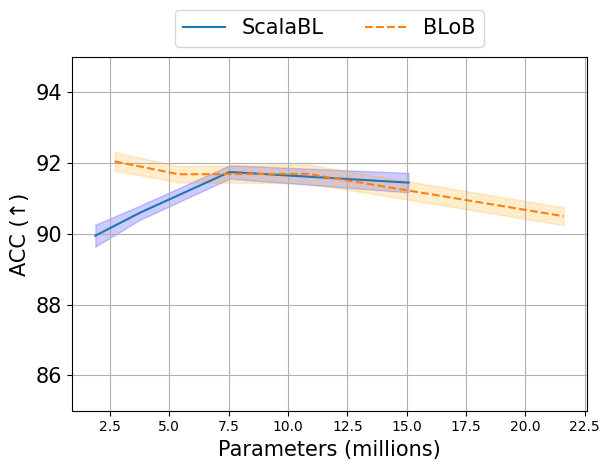}
  \end{minipage}\hfill
  \begin{minipage}{0.32\textwidth}\centering
    \includegraphics[width=\textwidth]{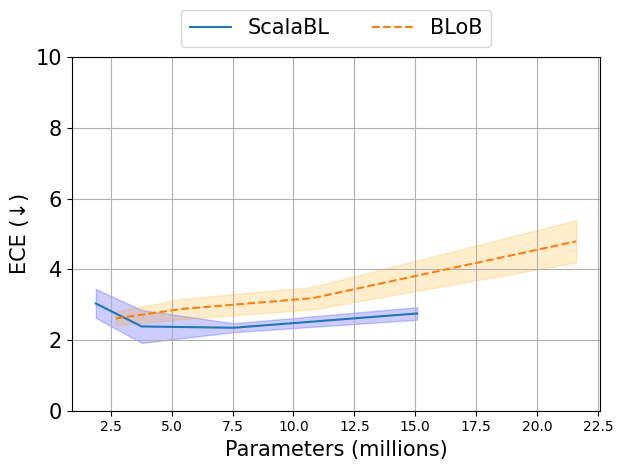}
  \end{minipage}\hfill
  \begin{minipage}{0.32\textwidth}\centering
    \includegraphics[width=\textwidth]{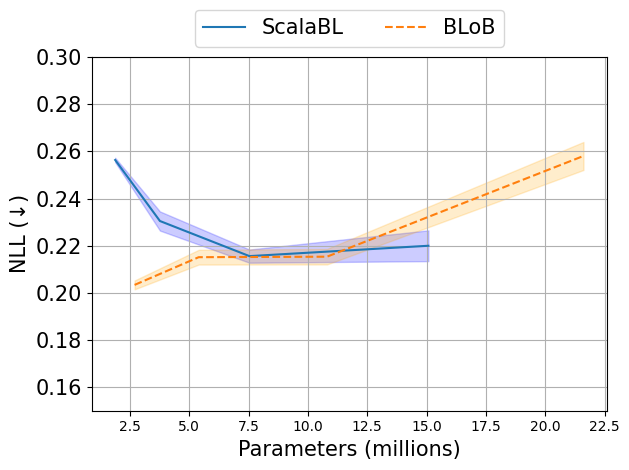}
  \end{minipage}\\[-0.5em]

  % ---------- BoolQ ----------
  \begin{minipage}{0.32\textwidth}\centering
    \includegraphics[width=\textwidth]{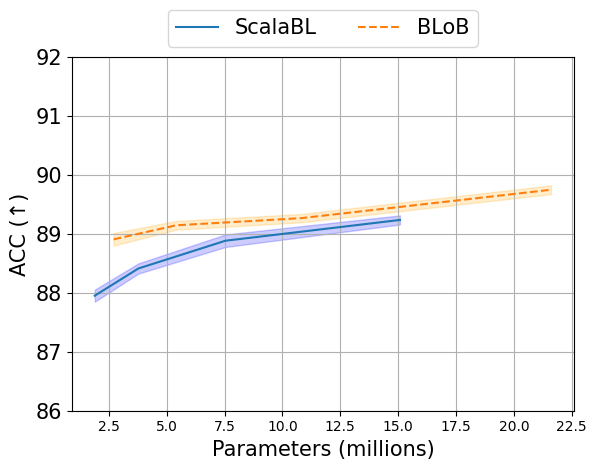}
  \end{minipage}\hfill
  \begin{minipage}{0.32\textwidth}\centering
    \includegraphics[width=\textwidth]{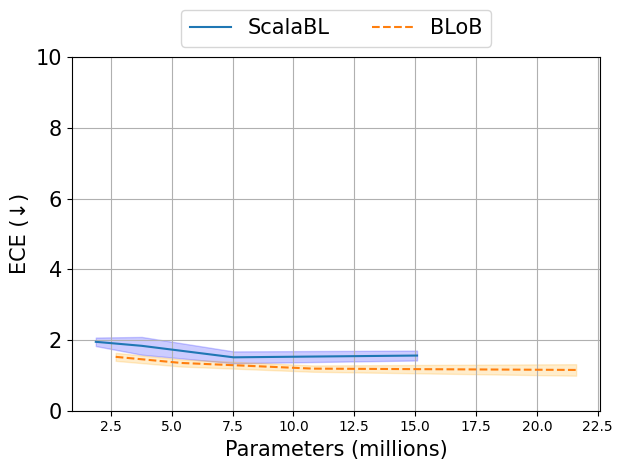}
  \end{minipage}\hfill
  \begin{minipage}{0.32\textwidth}\centering
    \includegraphics[width=\textwidth]{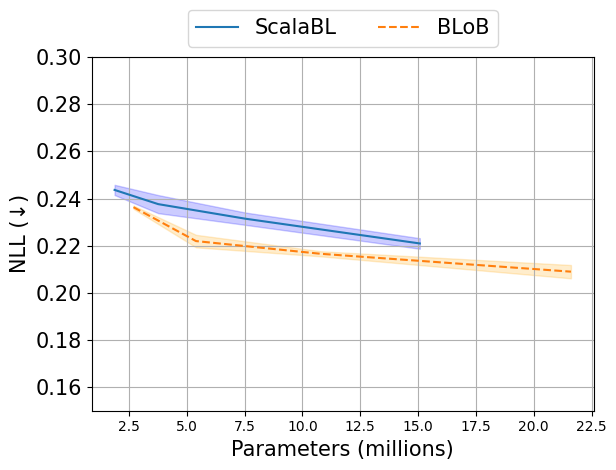}
  \end{minipage}

  \caption{Effect of LoRA rank $r$. \textbf{Top:} Winogrande-Medium (WG-M). \textbf{Middle:} OpenBookQA (OBQA). \textbf{Bottom:} BoolQ.}
  \label{fig:r_sweep2}
\end{figure*}